\documentclass[letterpaper]{article} % DO NOT CHANGE THIS
\usepackage[preprint]{aaai2027}  % DO NOT CHANGE THIS
\usepackage[hyphens]{url}  % DO NOT CHANGE THIS
\usepackage{graphicx} % DO NOT CHANGE THIS
\def\UrlFont{\rm}  % DO NOT CHANGE THIS
\usepackage{natbib}  % DO NOT CHANGE THIS AND DO NOT ADD ANY OPTIONS TO IT
\usepackage{caption} % DO NOT CHANGE THIS AND DO NOT ADD ANY OPTIONS TO IT
\usepackage{algorithm}
\usepackage{algorithmic}
\usepackage{amsmath,amssymb}
\usepackage{multirow}
\usepackage{booktabs}
\usepackage{makecell}
\usepackage{pifont}
\usepackage{xfp}
\usepackage[table]{xcolor} % 提供 \rowcolor
\definecolor{mmishade}{gray}{0.90}
\definecolor{textmoeshade}{gray}{0.80}
\usepackage{newfloat}
\usepackage{listings}
\DeclareCaptionStyle{ruled}{labelfont=normalfont,labelsep=colon,strut=off} % DO NOT CHANGE THIS
\floatstyle{ruled}
\newfloat{listing}{tb}{lst}{}
\floatname{listing}{Listing}

\usepackage{booktabs}
\usepackage{enumitem}

\makeatletter
\newsavebox{\tblbox}
\newcommand{\debugresizeboxfontsize}[2]{%
  \sbox{\tblbox}{#2}%
  \edef\basept{\f@size}%
  \edef\natpt{\strip@pt\wd\tblbox}%
  \edef\tgtpt{\strip@pt\dimexpr #1\relax}%
  \typeout{[TABLE] base=\basept pt, natural=\natpt pt, target=\tgtpt pt, scale=\fpeval{\tgtpt/\natpt}, effective=\fpeval{\basept*(\tgtpt/\natpt)} pt}%
  \resizebox{#1}{!}{\usebox{\tblbox}}%
}
\makeatother
\title{Semi-MedRef: Semi-Supervised Medical Referring Image Segmentation with Cross-Modal Alignment}
\author{
Yuchen Li\textsuperscript{\rm 1},
Ziru Wei\textsuperscript{\rm 2},
Zhen Zhao\textsuperscript{\rm 3},
Yi Liu\textsuperscript{\rm 4},
Luping Zhou\textsuperscript{\rm 1}
}

\affiliations{
\textsuperscript{\rm 1}The University of Sydney\\
\textsuperscript{\rm 2}The ATLAS Institute\\
\textsuperscript{\rm 3}Shanghai Artificial Intelligence Laboratory\\
\textsuperscript{\rm 4}Changzhou University
}

\begin{document}

\maketitle

\begin{abstract}

Medical referring image segmentation (MRIS) predicts lesion masks from medical images and natural-language referring expressions, but acquiring paired pixel-level annotations and referring texts is costly. Semi-supervised learning (SSL) can alleviate this burden by exploiting unlabeled data, yet its effectiveness depends on preserving image--text alignment under strong perturbations. Existing SSL methods for referring segmentation rely on independent or simple multimodal perturbations (\emph{e.g.,} left--right flips), while stronger augmentations such as CutMix remain largely unexplored because they can disrupt cross-modal correspondence. We propose Semi-MedRef, a teacher--student SSL framework that explicitly preserves alignment between medical images and positional language through three complementary components: T-PatchMix, an alignment-preserving cross-modal augmentation that synchronizes patch mixing with positional-language and pseudo-mask updates; PosAug, a position-aware text augmentation that regularizes reliance on positional expressions; and Positional Affinity Contrastive Learning (PACL), which exploits coarse positional cues to construct region-aware supervision through anatomically weighted soft positives, encouraging anatomically grounded cross-modal representation learning. Experiments on QaTa-COV19 and MosMedData+ demonstrate that Semi-MedRef consistently outperforms state-of-the-art fully supervised and semi-supervised MRIS methods across all label regimes.

\end{abstract}


\section{Introduction}
\begin{figure}[!t]
  \centering
  \includegraphics[width=\columnwidth]{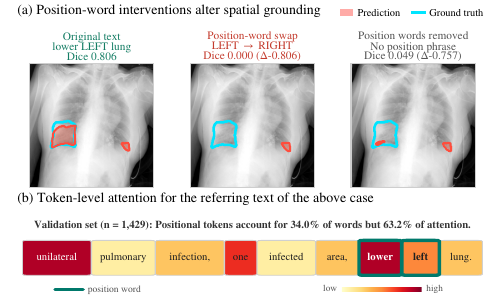}
%   \caption{Observations on position-aware cross-modal alignment.
% (a) Spatial grounding under controlled text interventions. 
% (b) Segmentation performance under positional masking, fuzzing, removal, directional swaps, and an \textbf{qual-count-balanced?} non-positional masking control.
% (c) Positional-token attention enrichment in the two deep fusion stages, normalized by token frequency.
% }
\caption{Controlled analysis of positional-language sensitivity in the MMI-UNet baseline. (a) Predictions for an example image under the original, laterality-swapped, and position-removed referring texts. (b) Token-level attention for the original referring text, with a validation-set summary of attention allocated to positional tokens.%where darker colors indicate larger attention weights. The validation-set statistic above the heatmap summarizes the relative attention enrichment of positional tokens.
}
\label{fig:motivation}
\end{figure}

Medical referring image segmentation (MRIS) predicts the pixel-level mask for one or more target lesions from a medical image and a natural-language referring expression. Unlike class-based segmentation, the expression specifies the intended target and often contains anatomical or relative spatial cues (\emph{e.g., left/right and upper/lower)} that help localize the referred lesion regions when multiple plausible regions are present. This text-conditioned formulation supports key clinical tasks such as lesion quantification, treatment planning, and disease monitoring~\cite{li2024lvit,bui2024visual,zhong2023ariadne}. 

However, obtaining paired pixel-level masks and referring expressions is costly, particularly for lesion-level tasks with ambiguous boundaries and high inter-observer variability. Semi-supervised learning (SSL)~\cite{bachman2014learning,berthelot2020remixmatch,zhao2022dc} mitigates this challenge by leveraging unlabeled data through pseudo-labeling and consistency regularization, assuming that perturbations preserve supervision semantics.

In recent dominant consistency-based studies, data augmentation is the primary mechanism for transferring supervision from the teacher's weak view to the student's strong view. Accordingly, the image, referring expression, and pseudo-mask must remain semantically consistent; otherwise, incorrect pseudo-supervision may exacerbate confirmation bias. This is especially critical in MRIS, where spatial transformations can change the meaning of positional expressions such as \emph{left}, \emph{right}, \emph{upper}, and \emph{lower}.

Existing SSL methods for general referring expression segmentation (RES)  provide only limited mechanisms to preserve such cross-modal consistency. RESMatch~\cite{zang2025resmatch} adjusts positional words under horizontal flips, SemiRES~\cite{yang2024sam} improves pseudo-label quality using SAM~\cite{kirillov2023segment}, and TextMatch~\cite{li2024textmatch} combines an EMA teacher with multimodal fusion and pseudo-label regularization. Because strong spatial perturbations require synchronized updates to visual
regions, position-sensitive textual spans, and teacher pseudo-masks, these methods largely rely on unimodal or weakly coupled transformations. Consequently, stronger augmentations, such as CutMix-style~\cite{yun2019cutmix,hong2023cmmix} patch replacement and random cropping, remain largely unexplored for multimodal teacher--student learning because they can disrupt image--text correspondence and corrupt pseudo-supervision. The problem is particularly acute in MRIS, where positional language provides essential spatial grounding: geometric transformations can invert left/right laterality, while lexical variation or partial masking can encourage brittle reliance on exact wording  rather than visual anatomical evidence. \textbf{These limitations expose a central challenge: how to preserve image--text alignment under strong perturbations without sacrificing positional semantics?}

Figure~\ref{fig:motivation} reveals the baseline's strong reliance on positional language. As shown in Figure~\ref{fig:motivation}(a), altering or removing positional expressions substantially redirects or degrades the predicted segmentation, while Figure~\ref{fig:motivation}(b) shows that positional tokens account for a disproportionate share of cross-modal attention. These observations suggest that positional language provides valuable spatial supervision but also induces brittle image--text correspondences under lexical perturbations. This motivates us to preserve positional consistency under strong perturbations, improve robustness to positional-language variations, and explicitly align positional language with visual evidence.

\begin{figure*}[t]
\centering
  \includegraphics[width=0.9\textwidth]{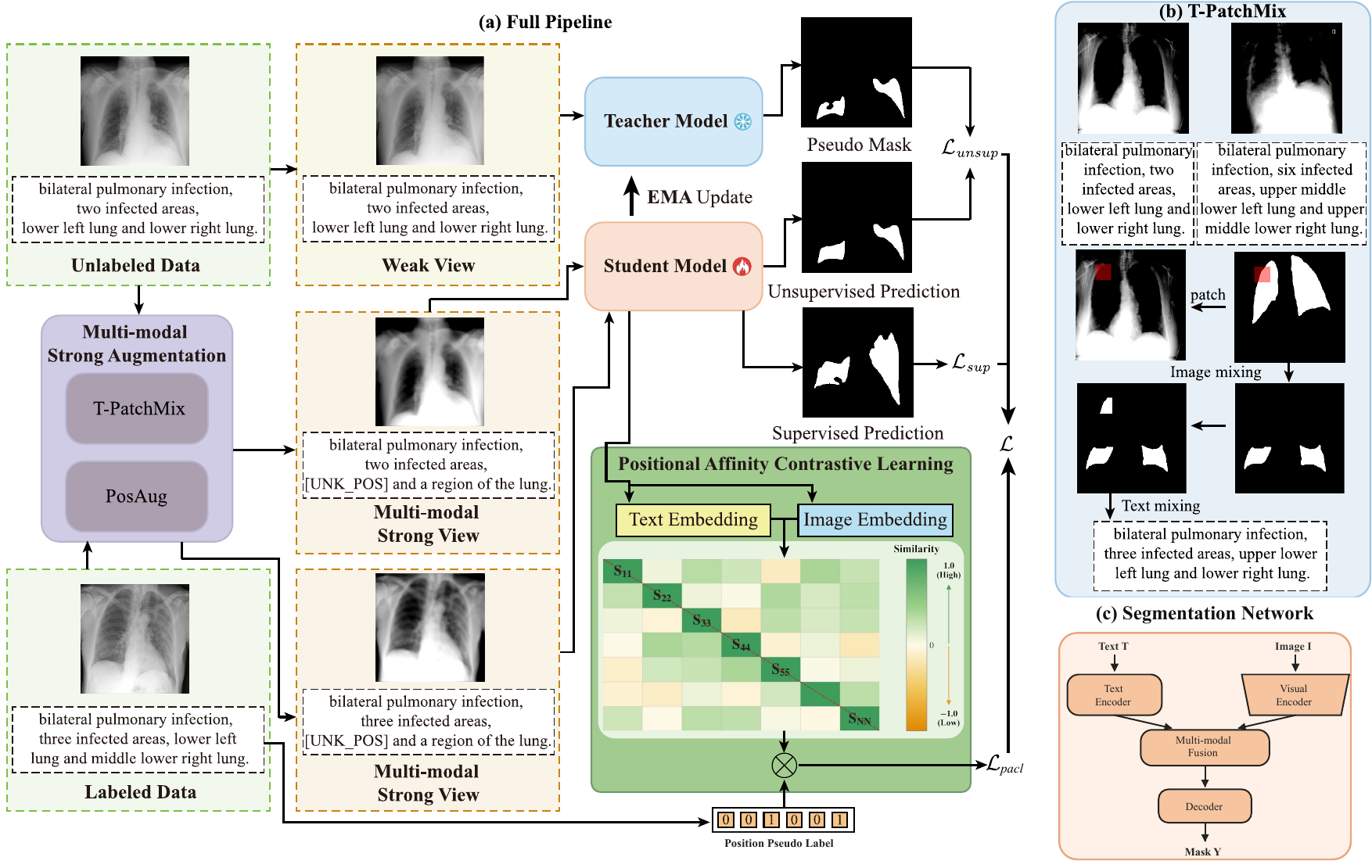}
  \caption{Overview of Semi-MedRef.
(a) The EMA teacher provides weak-view pseudo-masks, while the student
learns from strongly augmented views using PosAug, T-PatchMix, and PACL.
(b) T-PatchMix jointly transforms image patches, pseudo-masks, and
corresponding referring text spans to preserve cross-modal consistency.
(c) The text-conditioned segmentation network fuses textual and visual features to predict the referred lesion mask.} 
  \label{fig1}
\end{figure*}

Guided by these observations, we propose \textbf{Semi-MedRef} (Fig.~\ref{fig1}), a \textbf{Semi}-supervised \textbf{Med}ical \textbf{Ref}erring image segmentation framework that explicitly preserves cross-modal alignment between medical images and positional language under strong perturbations through three complementary components. T-PatchMix extends patch-level mixing to multimodal SSL by selectively mixing anatomically compatible regions while synchronously updating the corresponding positional spans, preserving spatial grounding under strong augmentation. PosAug regularizes the language stream by perturbing positional phrases, encouraging reliance on visual evidence when location cues are incomplete or imprecise. Positional Affinity Contrastive Learning (PACL) exploits coarse positional pseudo-labels to construct region-aware soft positives, strengthening anatomically grounded image--text alignment in the representation space. As an optional robustness extension for clinical reports with missing positional descriptions, Image-derived Soft Position Guidance (ISPG) converts image-predicted anatomical probabilities into a soft position token while retaining the available referring text.

In summary, our contributions are four-fold:
 
\begin{itemize}[nosep]
\item We propose Semi-MedRef, a semi-supervised MRIS framework that preserves cross-modal alignment under strong perturbations through alignment-aware augmentation and cross-modal supervision.
\item We introduce two alignment-preserving augmentation strategies: T-PatchMix synchronizes patch-level image mixing with positional-language and pseudo-mask updates, while PosAug regularizes positional expressions to improve robustness to positional-language variations.
\item We develop PACL for region-aware cross-modal representation learning using coarse positional cues, and introduce ISPG as an optional extension providing soft spatial guidance for texts with missing positional expressions.
\item Extensive experiments on two MRIS datasets, multiple backbone architectures, and the multi-domain PosMed benchmark demonstrate consistent improvements across all label regimes, validating the effectiveness and generalizability of Semi-MedRef.
%Extensive experiments on two MRIS datasets and multiple backbones demonstrate consistent improvements across all label regimes, validating the effectiveness and generality of Semi-MedRef.
\end{itemize}

\section{Related Work}
\paragraph{Semi-Supervised Referring Segmentation.}
Mean Teacher~\cite{tarvainen2017mean} and FixMatch~\cite{sohn2020fixmatch} establish teacher-based and weak-to-strong consistency, while CPS~\cite{chen2021semi}, U2PL~\cite{wang2022semi}, and UniMatch~\cite{yang2023revisiting,yang2025unimatch} extend these principles to semantic segmentation. Building on these paradigms, LAVT~\cite{yang2022lavt} studies language-aware fusion for referring segmentation, whereas SemiRES~\cite{yang2024sam} and RESMatch~\cite{zang2025resmatch} explore its semi-supervised setting. For MRIS, VLMs like LViT~\cite{li2024lvit}, GuideDecoder~\cite{zhong2023ariadne}, MMI-UNet~\cite{bui2024visual}, ViTexNet~\cite{bhardwaj2025vitexnet}, FMISeg~\cite{yu2025frequency}, and MG-UNet~\cite{ding2025mg} develop complementary image--text fusion architectures. Recent SSL-MRIS methods further exploit language: TextMatch~\cite{li2024textmatch} regularizes pseudo-labels with textual prompts; TextMoE~\cite{zeng2025exploring} uses text-enhanced experts; DuSSS~\cite{Pan_Qiao_Lou_Ji_Li_2025} and AMVLM~\cite{pan2025amvlm} model semantic similarity and alignment multiplicity, respectively; and
TeViD~\cite{zeng2026tevid} adopts text-anchored visual contrast. Related to incomplete language, SGSeg~\cite{10.1007/978-3-031-72111-3_23} derives coarse anatomical labels from reports and predicts them from images for text-free inference. Our ISPG instead retains the remaining referring text and uses the
image only to supply auxiliary positional guidance.

% TextMatch~\cite{li2024textmatch} incorporates textual prompts into semi-supervised segmentation, whereas TextMoE~\cite{zeng2025exploring} learns from labeled and unlabeled image--text data using a text-assisted mixture-of-experts framework. Existing medical vision--language models instead mainly study fusion architectures: GuideDecoder~\cite{zhong2023ariadne} and MMI-UNet~\cite{bui2024visual} fuse reports with visual features, ViTexNet~\cite{bhardwaj2025vitexnet} employs text-guided dynamic convolution, FMISeg~\cite{yu2025frequency} performs frequency-domain fusion, and MG-UNet~\cite{ding2025mg} transfers textual knowledge into a memory bank. In contrast, Semi-MedRef addresses how image--text correspondence should be preserved when strong perturbations define the supervision transferred to unlabeled samples.

\paragraph{Multimodal Augmentation and Alignment.}
% Unlike image-only augmentation, multimodal augmentation must increase
% data diversity while preserving the semantic correspondence between
% visual content and language. ITMix~\cite{hong2022itmix} and
% MixGen~\cite{hao2023mixgen} mix paired image--text samples for
% vision--language representation learning. More closely related to
% grounding, CMMix~\cite{hong2023cmmix} jointly mixes visual regions and
% their textual descriptions, while Augment the
% Pairs~\cite{yi2024augment} distinguishes text-conditioned and
% text-independent transformations to maintain phrase--region
% consistency. In semi-supervised RES, RESMatch~\cite{zang2025resmatch}
% combines visual and textual perturbations with transformation-specific
% spatial-word correction, whereas SemiRES~\cite{yang2024sam} primarily
% improves pseudo-mask quality using SAM. In the medical domain,
% TextMatch~\cite{li2024textmatch} incorporates text prompts into
% semi-supervised segmentation but does not explicitly formulate strong
% paired perturbations for referring expressions. In contrast,
% Semi-MedRef targets the cross-modal inconsistency introduced by
% patch-level mixing: T-PatchMix constrains patch selection using
% anatomical compatibility and teacher confidence while synchronously
% updating positional phrases, PosAug regularizes the positional language
% stream, and ITCL reinforces region-aware image--text alignment.

Image-only methods such as CutMix~\cite{yun2019cutmix} and ClassMix~\cite{olsson2021classmix} perturb visual inputs without considering paired language. Multimodal approaches, including
ITMix~\cite{hong2022itmix}, MixGen~\cite{hao2023mixgen}, CMMix~\cite{hong2023cmmix}, and Augment the Pairs~\cite{yi2024augment}, jointly transform image--text pairs for representation learning or visual grounding. However, teacher--student MRIS additionally requires the referring expression and pseudo-mask to remain consistent with the transformed image. RESMatch corrects spatial words under simple geometric operations but does not support anatomy-aware patch-level mixing. Our Semi-MedRef synchronizes image patches, positional spans, and pseudo-masks through T-PatchMix, regularizes lexical variations with PosAug, and models graded anatomical compatibility with PACL.

% Image-only methods such as CutMix~\cite{yun2019cutmix},
% ClassMix~\cite{olsson2021classmix}, and UniMatch perturb visual inputs without considering paired language. Multi-modal approaches, including ITMix~\cite{hong2022itmix}, MixGen~\cite{hao2023mixgen}, CMMix~\cite{hong2023cmmix}, and Augment the Pairs~\cite{yi2024augment}, jointly modify images and text for representation learning or visual grounding. However, they are not designed for teacher--student MRIS, where the image, referring expression, and pseudo-mask must undergo consistent transformations. RESMatch corrects spatial words under simple geometric operations, but does not support anatomy-aware patch-level mixing. Semi-MedRef therefore introduces T-PatchMix to synchronously transform image regions, positional phrases, and pseudo-labels, PosAug to regularize positional language, and PACL to explicitly reinforce anatomical cross-modal alignment.

\section{Method}
\subsection{Problem Setting}
In semi-supervised MRIS task, each input is an image--text pair $(I,T)$, and the model predicts a binary mask
$\hat{Y}\in\{0,1\}^{H\times W}$.
The training data include a small labeled set $\mathcal{D}_{\ell}$ and a much larger unlabeled set $\mathcal{D}_u$:
% \begin{equation}
% \mathcal{D}_{\ell} = \{((I_i^{l}, T_i^{l}), Y_i^{l})\}_{i=1}^{N_l}, \qquad
% \mathcal{D}_u = \{(I_j^{u}, T_j^{u})\}_{j=1}^{N_u}, \qquad N_l \ll N_u,
% \end{equation}
\begin{equation}
\begin{aligned}
\mathcal{D}_{\ell}
&= \{((I_i^{\ell}, T_i^{\ell}), Y_i^{\ell})\}_{i=1}^{N_{\ell}}, \\
\mathcal{D}_{u}
&= \{(I_j^{u}, T_j^{u})\}_{j=1}^{N_u},
\quad N_{\ell} \ll N_u .
\end{aligned}
\end{equation}
where $(I_i^{l}, T_i^{l})$ and $(I_j^{u}, T_j^{u})$ denote labeled and unlabeled image--text pairs, respectively.
For unlabeled samples, no ground-truth masks are available. 
We learn a multimodal segmentation model $f_{\theta}$ that outputs a pixel-wise probability map $P=\sigma(f_{\theta}(I, T))\in[0,1]^{H\times W}$ after sigmoid $\sigma(\cdot)$.
% \begin{equation}
% P=\sigma(f_{\theta}(I, T))\in[0,1]^{H\times W}.
% \end{equation}
% where $P$ is the corresponding probability map after sigmoid function $\sigma(\cdot)$.

\subsection{Overview}
\label{sec:ssl}
% \subsection{Semi-supervised Teacher--Student Learning}
% \label{sec:ssl}

% Our method is a backbone-agnostic framework; any referring segmentation architecture with image--text fusion can be plugged in.
% In this work, we instantiate it with MMI-UNet~\cite{bui2024visual}.
% We adopt an EMA~\cite{tarvainen2017mean} teacher--student framework~\cite{sohn2020fixmatch} and optimize it in two phases: (i) a labeled-only burn-in for stable initialization~\cite{liu2021unbiased,liu2022unbiased,mi2022active}, and (ii) semi-supervised learning where the teacher generates pseudo-masks on weakly augmented unlabeled samples and the student learns under strong augmentation.
Fig.~\ref{fig1} illustrates Semi-MedRef, our teacher–student framework for SSL-based MRIS. The student receives strongly augmented image–text pairs, combining T-PatchMix for alignment-preserving patch mixing and PosAug for position-aware text perturbation, while the teacher processes weak views to provide stable segmentation pseudo-labels. Coarse positional pseudo-labels extracted from the referring text serve as soft positives for PACL, a structured contrastive objective that preserves graded referential compatibility. The student is trained with supervised loss, consistency loss, and PACL, enabling Semi-MedRef to maintain cross-modal coherence under strong perturbations. 
% In this work, we instantiate MedRef with MMI-UNet~\cite{bui2024visual}, though our strategies are compatible with other MRIS backbones, as demonstrated in Table~\ref{tab:qata_main}.

We optimize Semi-MedRef in two phases: (i) the labeled-only burn-in stage for stable initialization~\cite{liu2022unbiased,mi2022active}, and (ii) the semi-supervised learning stage where the teacher generates pseudo-masks on weakly augmented unlabeled samples and the student learns under strong augmentation.
In the burn-in stage, we train the student model on $\mathcal{D}_{\ell}$ with:
\begin{equation}
\mathcal{L}_{sup}
=
\mathcal{L}_{\text{DiceCE}}\!\left(\sigma\!\left(f_{\theta^{S}}(I_i^{l},T_i^{l})\right),\, Y_i^{l}\right),
\label{eq:sup_burnin_simplified}
\end{equation}
where $\mathcal{L}_{\text{DiceCE}}$~\cite{bui2024visual} is the Dice loss~\cite{7785132} plus pixel-wise binary cross-entropy~\cite{10.1007/978-3-319-24574-4_28}.

In the semi-supervised learning stage, we update the teacher parameters by EMA: $\theta^{T}\leftarrow m\,\theta^{T}+(1-m)\,\theta^{S}$, with decay coefficient $m=0.999$.
Given an unlabeled pair $(I_i^{u},T_i^{u})\in\mathcal{D}_u$, the teacher predicts on the weak view $(I_{i,w}^{u},T_{i,w}^{u})$ and produces a hard pseudo-mask $\tilde{Y}_i^{u}$ with confidence threshold $\delta$:
\begin{equation}
P_{i,w}^{T}=\sigma\!\left(f_{\theta^{T}}(I_{i,w}^{u},T_{i,w}^{u})\right),
\qquad
\tilde{Y}_i^{u}=\mathbb{I}\!\left(P_{i,w}^{T}\ge \delta\right).
\label{eq:pseudo_labeling_simplified}
\end{equation}
The student is trained on the corresponding strong view $(I_{i,s}^{u},T_{i,s}^{u})$ to match $\tilde{Y}_i^{u}$:
\begin{equation}
\mathcal{L}_{unsup}
=
\mathcal{L}_{\text{DiceCE}}\!\left(\sigma\!\left(f_{\theta^{S}}(I_{i,s}^{u},T_{i,s}^{u})\right),\,\tilde{Y}_i^{u}\right).
\label{eq:unsup_simplified}
\end{equation}
Finally, we optimize the student with the overall objective:
\begin{equation}
\mathcal{L}
=
\mathcal{L}_{sup}
+
\lambda_u\,\mathcal{L}_{unsup}
+
\lambda_{pacl}^{sup}\mathcal{L}_{pacl}^{sup}
+
\lambda_{pacl}^{unsup}\mathcal{L}_{pacl}^{unsup}.
\label{eq:loss_total}
\end{equation}
Here, $\lambda_u$ weights the unsupervised term, while $\lambda_{pacl}^{sup}$ and $\lambda_{pacl}^{unsup}$ weight $\mathcal{L}_{pacl}$ on labeled and unlabeled branches, respectively.

\subsection{Multi-modal Augmentation}
\label{sec:aug}

To preserve image--text alignment under strong perturbations, we design MRIS-tailored text and multimodal augmentations.

\paragraph{Position-aware Text Augmentation (PosAug).}
Positional phrases (\emph{e.g.,} \textit{upper left lung}) provide critical grounding cues but can be sensitive to wording variations under limited supervision~\cite{10.1007/978-3-031-72111-3_23}.
On the student branch, we extract and normalize the positional phrase, and then stochastically apply one of two edits:
(i) \emph{dropout}, which replaces the phrase with \texttt{[UNK\_POS]}; or
(ii) \emph{fuzzing}, which substitutes it with a location-agnostic phrase (\emph{e.g.,} \textit{upper left lung} $\rightarrow$ \textit{a region of the lung}).
Dropout is applied with probability $\rho$, and fuzzing with probability $1-\rho$.
The teacher retains the complete text to provide stable pseudo-labels, forming a weak-to-strong consistency task under incomplete positional cues.

\paragraph{Image--Text Alignment-preserving Patch Mixing (T-PatchMix).}
To benefit from mixing while preserving referring alignment, as illustrated in Fig.~\ref{fig1}, we construct the student strong view by patch-mixing~\cite{hong2024patchmix} two unlabeled pairs
$(I_i^{u},T_i^{u})$ and $(I_j^{u},T_j^{u})$ with a block-wise binary mask $M_i$:
\begin{equation}
I_{mix} = (1-M_i)\odot I_i^{u} + M_i\odot I_j^{u}.
\label{eq:xpatchmix_img}
\end{equation}
We consider three sampling variants:
(i) \emph{T-PatchMix-pos} extracts a positional phrase from the student text $T_j^{u}$, maps it to a coarse region
(left/right $\times$ upper/middle/lower), and samples $M_i$ only within this region; mixing is disabled when two texts conflict in laterality.
(ii) \emph{T-PatchMix-prob} samples $M_i$ from high-confidence regions of the teacher probability map $P_{j,w}^{T}$ on weak view of sample $j$. 
(iii) \emph{T-PatchMix-hybrid} applies both constraints by sampling a teacher-positive patch center within the region derived from \(T_j^u\).
To avoid mixing background or non-lesion patches, we define a lesion-gating ratio $r_i$ measuring the lesion fraction inside the mixed patch:
\begin{equation}
r_i=\frac{\sum (P_{j,w}^{T}\odot M_i)}{\sum M_i+\epsilon},\qquad
\hat{M}_i = M_i\cdot \mathbb{I}(r_i\ge \delta).
\label{eq:xpatchmix_gate}
\end{equation}
Here, $\delta$ is the same threshold in Eq.~\ref{eq:pseudo_labeling_simplified}. The mixed pseudo-mask is obtained by combining teacher probabilities:
\begin{equation}
\tilde{Y}_{\text{mix}}^{(i)}
=
\mathbb{I}\!\left(\left[(1-\hat{M}_i)\odot P_{i,w}^{T}+\hat{M}_i\odot P_{j,w}^{T}\right]\ge 0.5\right).
\label{eq:xpatchmix_pseudo}
\end{equation}
We synchronize the text with the mixed image to maintain semantic consistency. We apply position-aware span mixing (\emph{e.g.,} “pos$_i$ and pos$_j$”) only when $\hat{M}_i\neq 0$. Span mixing replaces or conjoins positional phrases and synchronously updates coupled attributes, such as lesion count and laterality.

% We synchronize the text with the mixed image to maintain semantic consistency by applying alignment-aware span mixing when $\hat{M}_i \neq 0$. The positional phrases are transferred or composed (\emph{e.g.,} ``pos$_i$ and pos$_j$''), while coupled attributes, including lesion count, laterality, and applicable donor descriptors, are updated according to the mixed content. Otherwise, the original caption is retained.

% We synchronize the text with the mixed image to maintain semantic consistency. When $\hat{M}_i \neq 0$, we apply alignment-aware span mixing to all target-related textual attributes. Specifically, the positional phrase from the donor caption is transferred to or composed with that of the recipient caption (\emph{e.g.,} ``pos$_i$ and pos$_j$''), while semantically coupled attributes, such as lesion count and laterality, are synchronously updated to reflect the mixed content. Other target-related descriptors associated with the donor patch are transferred together when applicable. Therefore, the mixed image, pseudo-mask, and all affected textual spans remain mutually consistent. When $\hat{M}_i = 0$, the original image, pseudo-mask, and caption are retained.

\begin{table*}[t]
\centering

{\small
\setlength{\tabcolsep}{3.5pt}
\renewcommand{\arraystretch}{1.0}

\begin{tabular}{@{}l*{8}{c}@{}}
\toprule
\multirow{2}{*}{Method}
& \multicolumn{4}{c}{\textbf{QaTa-COV19}}
& \multicolumn{4}{c}{\textbf{MosMedData+}} \\
\cmidrule(lr){2-5}
\cmidrule(lr){6-9}
& 2\% & 5\% & 15\% & 100\%
& 2\% & 5\% & 15\% & 100\% \\
\midrule

LAVT (CVPR\,\textquotesingle22)
& 69.25/57.56
& 73.93/63.24
& 76.61/66.49
& 79.28/69.89
& 58.44/40.23
& 67.54/55.39
& 69.75/57.64
& 73.29/60.41 \\

SemiRES (ICML\,\textquotesingle24)
& 75.33/64.82
& 77.40/67.45
& 77.52/67.59
& 80.16/70.20
& 61.90/46.22
& 68.64/55.96
& 71.03/59.07
& 74.40/61.05 \\

\midrule

ViTexNet (MICCAI\,\textquotesingle25)
& 62.08/45.01
& 75.42/60.54
& 88.29/79.04
& 89.71/81.35
& 62.34/45.28
& 67.90/52.13
& 71.11/55.62
& 77.39/63.46 \\

FMISeg (MICCAI\,\textquotesingle25)
& 83.99/72.40
& 86.49/76.17
& 88.07/78.68
& 90.82/83.19
& 63.49/46.51
& 65.65/48.87
& 73.68/58.35
& 76.30/61.44 \\

MG-UNet (MICCAI\,\textquotesingle25)
& 83.13/71.13
& 86.48/76.19
& 88.02/78.60
& 90.76/83.09
& 59.53/42.38
& 67.81/51.30
& 71.37/55.48 
& 76.39/61.79 \\

GuideDecoder (MICCAI\,\textquotesingle23)
& 80.44/67.28
& 84.23/72.76
& 85.03/73.95
& 89.78/81.45
& 66.34/49.63
& 71.21/55.30
& 73.10/57.95
& 77.75/63.60 \\

\rowcolor{mmishade}
MMI-UNet (MICCAI\,\textquotesingle24)
& 84.63/73.36
& 87.35/77.55
& 88.72/79.73
& 90.88/83.28
& 68.23/51.78
& 72.26/56.57
& 74.00/58.73
& 78.42/64.50 \\

\midrule

LViT (TMI\,\textquotesingle24)
& 75.23/64.51
& 77.24/66.31
& 79.98/70.04
& 83.66/75.11
& 64.19/52.19
& 68.61/56.49
& 71.45/59.64
& 74.57/61.33 \\

TextMatch (MICCAI\,\textquotesingle24)
& 80.42/67.25
& 83.56/71.67
& 86.21/75.64
& 89.15/81.75
& 62.73/45.70
& 72.46/58.12
& 75.73/60.93
& 76.21/61.88 \\

DuSSS (AAAI\,\textquotesingle25)
& 61.88/48.84
& 71.73/60.51
& 73.93/62.85
& 87.15/77.23
& 50.01/37.31
& 53.81/41.34
& 57.22/45.28
& 73.17/59.83 \\

AMVLM (TMI\,\textquotesingle25)
& 64.93/48.07
& 73.12/57.63
& 78.26/64.28
& 87.28/77.43
& 46.89/30.63
& 55.27/38.19
& 61.81/44.67
& 74.16/58.94 \\

\rowcolor{textmoeshade}
TextMoE (MICCAI\,\textquotesingle25)
& 83.58/71.79
& 85.47/74.62
& 87.75/78.18
& 90.84/83.21
& 63.23/46.24
& 71.84/56.05
& 73.91/58.61
& 75.68/61.51\\

TeViD (TIP\,\textquotesingle26)
& 84.50/73.15
& 86.10/75.59
& 88.15/78.82
& 90.51/82.69
& 66.99/50.37
& 72.15/56.43
& 75.18/60.23
& 77.60/63.40 \\

\midrule

% \rowcolor{gray!18}
% \textbf{Ours (GuideDecoder)}
% & 86.07/75.54
% & 87.12/77.18
% & 89.13/80.39
% & 90.91/83.34
% & \textbf{72.87/57.32}
% & 73.11/57.62
% & 74.70/59.61
% & 77.94/63.86 \\

% \textbf{Ours (MMI-UNet)}
% & $86.87{\scriptstyle\pm0.03}/76.79{\scriptstyle\pm0.05}$
% & $88.36{\scriptstyle\pm0.34}/79.14{\scriptstyle\pm0.54}$
% & $89.51{\scriptstyle\pm0.20}/81.01{\scriptstyle\pm0.32}$
% & $90.93{\scriptstyle\pm0.16}/83.37{\scriptstyle\pm0.26}$
% & $\mathbf{73.06{\scriptstyle\pm0.76}/57.56{\scriptstyle\pm0.94}}$
% & $73.43{\scriptstyle\pm0.90}/58.02{\scriptstyle\pm1.12}$
% & $76.26{\scriptstyle\pm0.29}/61.63{\scriptstyle\pm0.38}$
% & $77.87{\scriptstyle\pm0.26}/63.76{\scriptstyle\pm0.36}$ \\

\rowcolor{mmishade}
\textbf{Ours (MMI-UNet)}
& 87.25/77.38
& 88.31/78.98
& 89.84/81.56
& \textbf{91.39/84.14}
& \textbf{73.06/57.56}
& \textbf{74.51/59.38}
& \textbf{76.42/61.84}
& \textbf{78.70/64.89} \\

\rowcolor{mmishade}
\quad\textit{$\pm$ Std.}
& 0.03/0.05
& 0.34/0.54
& 0.20/0.32
& 0.16/0.26
& 0.76/0.94
& 0.90/1.12
& 0.29/0.38
& 0.26/0.36 \\

% \rowcolor{gray!18}
% \textbf{Ours (ViTexNet)}
% & 86.34/75.97
% & 87.94/78.47
% & 88.66/79.63
% & 90.25/82.24
% & 69.66/53.44
% & 71.94/56.17
% & 73.88/58.58
% & 76.24/61.61 \\

% \rowcolor{gray!18}
% \textbf{Ours (FMISeg)}
% & 85.77/75.04
% & 87.49/77.58
% & 89.37/80.77
% & 91.18/83.30
% & 70.12/51.91
% & 70.60/54.80
% & 73.46/58.22
% & 76.29/59.88 \\

% \rowcolor{gray!18}
% \textbf{Ours (MG-UNet)}
% & 87.03/77.03
% & 87.94/78.48
% & 89.05/80.26
% & 90.77/83.11
% & 72.52/56.89
% & 73.07/57.57
% & 74.20/58.98
% & 75.64/60.83 \\

%%%%%%%上面是可以放在附录里的实验结果

\rowcolor{textmoeshade}
\textbf{Ours (TextMoE-Seg)}$^{\ddagger}$
& \textbf{87.63/77.98}
& \textbf{88.68/79.66}
& \textbf{90.41/82.50}
& 91.31/84.01
& 71.72/55.91
& 72.57/56.72
& 75.83/61.07
& 77.65/63.47 \\

\rowcolor{textmoeshade}
\quad\textit{$\pm$ Std.}
& 0.23/0.35
& 0.22/0.35
& 0.41/0.66
& 0.12/0.19
& 0.98/1.18
& 0.51/0.64
& 0.28/0.36
& 0.71/0.94 \\

% \rowcolor{gray!18}
% \textbf{Ours (TeViD)}
% & 87.61/77.97
% & 88.14/78.80
% & 89.76/81.41
% & 91.31/84.02
% & 71.72/55.94
% & 74.77/59.72
% & 76.30/61.68
% & 78.13/64.13 \\
%%%%%可以放在附录的实验结果

% \rowcolor{gray!18}
% \quad std.
% & 0.03/0.05 & 0.34/0.54 & ... \\

\bottomrule
\end{tabular}
}

\caption{Performance comparison (Dice/mIoU (\%), higher is better) on QaTa-COV19 and MosMedData+ under different labeled-data ratios. Shaded rows pair each backbone with its corresponding Semi-MedRef variant. $^{\ddagger}$TextMoE-Seg retains TextMoE's report-conditioned MoE segmentation network while replacing its native SSL training with Semi-MedRef.
%Performance comparison on QaTa-COV19 and MosMedData+
%under different label ratios. Each entry reports Dice/mIoU (\%). Matched shading links each native architecture to its Semi-MedRef instantiation. $^{\ddagger}$For TextMoE-Seg, we retain TextMoE's text-conditioned MoE segmentation network but replace its native SSL training with Semi-MedRef. The
%following row reports Dice/mIoU standard deviations over three
%random seeds. Higher is better. 
}
\label{tab:qata_main}
\end{table*}
\subsection{Positional Affinity Contrastive Learning (PACL)}
\label{sec:pacl}
Standard CLIP~\cite{radford2021learning} loss treats all unmatched referring texts as equally negative. In MRIS, however, unmatched referring texts may describe partially overlapping sets of anatomical regions (e.g., left-upper and left-middle vs. left-middle and left-lower lung involvement), retaining substantial positional compatibility. We therefore introduce PACL, which replaces binary pairwise supervision with graded anatomical-affinity weights derived from coarse positional pseudo-labels.
As shown in Fig.~\ref{fig1}, for an image--text pair $(I,T)$, the visual encoder produces a pre-fusion image feature map $F$.
We obtain an image embedding $v=\mathrm{Proj}_I(\mathrm{GAP}(F))\in\mathbb{R}^{d}$ and a text embedding
$u=\mathrm{Proj}_T(T)\in\mathbb{R}^{d}$, and $\ell_2$-normalize both.
For a mini-batch of size $B$, we compute similarities $S_{ij}=v_i^\top u_j/\tau$, 
where $\tau$ is the temperature. To construct soft positives, we use a position pseudo-label $q_i\in\{0,1\}^{6}$ encoding the coarse anatomical regions (left/right $\times$ upper/middle/lower) described by the text. This coarse representation is intentional rather than an approximation of the segmentation mask. 

Clinical referring expressions typically specify approximate laterality and vertical location but do not provide reliable pixel-level coordinates. A finer spatial encoding would therefore impose unsupported precision and risk turning the auxiliary cue into a surrogate segmentation label. In contrast, the multi-hot left/right \(\times\) upper/middle/lower representation is robust to lexical variation, accommodates bilateral and multi-region findings, and provides the anatomical compatibility required by PACL without replacing pixel-level segmentation supervision. Then we compute the Jaccard affinity as $A_{ij}=|q_i\cap q_j|/|q_i\cup q_j|$ with $A_{ii}=1$, and optimize a bidirectional weighted contrastive loss:
\begin{equation}
\label{eq:itcl_loss_compact}
\begin{aligned}
\mathcal{L}_{pacl}
&=
-\frac{1}{2B}\sum_{i=1}^{B}\Bigg(
\frac{\sum_{j}A_{ij}\log \mathrm{softmax}(S_{ij})}{\sum_{j}A_{ij}} \\
&\qquad\qquad\qquad\quad
+
\frac{\sum_{j}A_{ij}\log \mathrm{softmax}(S_{ji})}{\sum_{j}A_{ij}}
\Bigg).
\end{aligned}
\end{equation}

\subsection{Optional Robustness Extension: Image-derived Soft Position Guidance}

Reusing the six-region position label $q_i$ defined above,
ISPG predicts an image-derived soft position token:
\begin{equation}
\begin{gathered}
\hat{q}_i=\sigma\bigl(g_{\phi}(I_i)\bigr),\qquad
\mathbf z_i^{p}=h_{\psi}\bigl(\hat{q}_i\bigr),\\
\hat{Y}^{s}_{i}
=f_{\theta_S}\bigl(
I_i^{s},A_{\rho_{\mathrm{tr}}}(T_i),\mathbf z_i^{p}
\bigr).
\end{gathered}
\label{eq:ispg}
\end{equation}
Here, $A_{\rho_{\mathrm{tr}}}$ removes all positional spans from a
referring text with probability $\rho_{\mathrm{tr}}$ while retaining the
remaining text. Report-derived labels supervise $g_{\phi}$ on both
segmentation-labeled and unlabeled images using a BCE-based multi-label
loss; the two losses are added to $\mathcal L$ with weight $0.1$ each
and require no additional annotation. The soft token is appended to
the language sequence and used in all fusion stages. During
removal-aware training, the EMA teacher uses the complete referring text with
$q_i$, whereas the student follows Eq.~\eqref{eq:ispg}.
At inference, $\hat{q}_i$ supplies the token when positional
expressions are absent, while the remaining referring text is retained.

\section{Experiments and Results}
%\subsection{Datasets, Metrics and Implementation}
\subsection{Datasets and Metrics.}
We evaluate Semi-MedRef on two public MRIS datasets, 
% QaTa-COV19~\cite{degerli2022osegnet,10.1007/978-3-031-72111-3_23} and MosMedData+~\cite{hofmanninger2020automatic,morozov2020mosmeddata}, 
QaTa-COV19~\cite{degerli2022osegnet} and MosMedData+~\cite{morozov2020mosmeddata},
following the official splits. 
QaTa-COV19 comprises 5{,}716 training, 1{,}429 validation, and 2{,}113 test samples. MosMedData+ comprises 2{,}183 training, 273 validation, and 273 test samples.
%The referring texts are structured infection-site descriptions from prior work~\cite{li2024lvit}.
The referring texts are structured infection-site descriptions from prior work~\cite{li2024lvit}. For semi-supervised learning, we randomly sample labeled subsets with ratios $r\in\{2\%,5\%,15\%\}$ from the training split, treating the remaining samples as unlabeled. The six-region positional pseudo-labels used by PACL and ISPG require no additional annotation: we use the report-derived metadata released for QaTa-COV19 by SGSeg~\cite{10.1007/978-3-031-72111-3_23} and deterministically extract equivalent tags from the MosMedData+ referring expressions. These labels are used only during training and not required for segmentation inference. Performance is evaluated using Dice and mIoU.
%For semi-supervised learning, we randomly sample labeled subsets from the training split with ratios $r\in\{2\%,5\%,15\%\}$ and treat the remaining training samples as unlabeled. The six-region positional pseudo-labels used by PACL and ISPG require no additional annotation: we use the report-derived metadata released for QaTa-COV19 by SGSeg~\cite{10.1007/978-3-031-72111-3_23}, and deterministically extract equivalent tags from the existing MosMedData+ referring expressions. They are used only during training and are not required for segmentation inference. We employ Dice and mIoU to evaluate the performance.
\subsection{Implementation Details.}
We implement our method in PyTorch Lightning and MONAI~\cite{cardoso2022monai}. Each training run uses a single NVIDIA GeForce RTX 3090 GPU with 24 GB memory. Input images are resized to $224\times224$. Weak augmentation uses RandomZoom. Image strong augmentation applies ColorJitter and GaussianBlur. We optimize with AdamW~\cite{loshchilov2019decoupled} with an initial learning rate of $3\times10^{-4}$, cosine annealing of $10^{-6}$, and a batch size of 32. 
% The EMA coefficient is warmed up from $m_{\text{start}}=0.99$ to $m_{\text{end}}=0.999$ over 20 epochs. We use a burn-in of 5 epochs, ramp up the unsupervised weight $\lambda_u$ to 0.3 over 15 epochs, and set the pseudo-label confidence threshold to $\delta=0.5$.
For PACL, we set $\tau=0.07$, $\lambda_{pacl}^{sup}=0.02$, and $\lambda_{pacl}^{unsup}=0.1$. For multimodal augmentation, PosAug uses $\rho=0.5$ and T-PatchMix uses a block size of 16. Unless noted, we use MMI-UNet~\cite{bui2024visual} with a ConvNeXt~\cite{liu2022convnet} visual encoder and a CXR-BERT~\cite{boecking2022making} text encoder. ISPG is used only for position-degraded robustness experiments. More implementation details are provided in the supplementary material.
%For PACL, we set temperature $\tau$ as 0.07, supervised weight $\lambda_{pacl}^{sup}$ as 0.02 and unsupervised weight $\lambda_{pacl}^{unsup}$ as 0.1. For multimodal augmentation, PosAug dropout probability $\rho$ is 0.5 and block size of T-PatchMix is $16$. Unless noted, we use MMI-UNet~\cite{bui2024visual} with a ConvNeXt~\cite{liu2022convnet} visual encoder and a CXR-BERT~\cite{boecking2022making} text encoder. ISPG is activated only
%in the position-degraded robustness experiments. More details are provided in supplementary material.
%---------------------%
\subsection{Comparison with State-of-the-Art Methods}
\label{sec:perf_comp}
\begin{figure*}[t]
  \centering
  \includegraphics[width=0.95\textwidth]{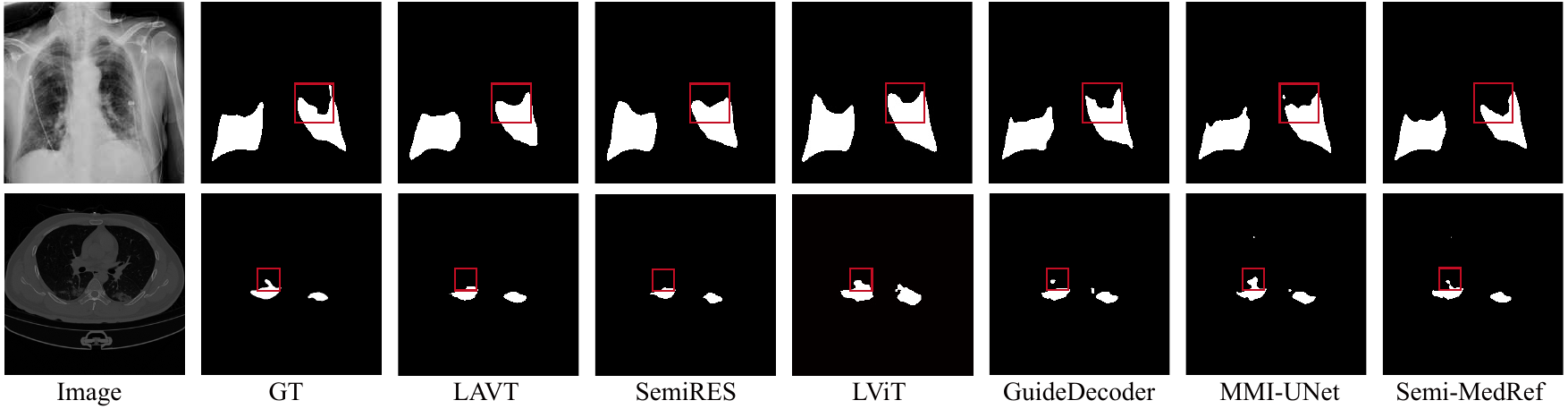}
  % \captionsetup{font=footnotesize, skip=5pt}
  \caption{Qualitative results on QaTa-COV19 (top) and MosMedData+ (bottom); red boxes mark boundary/small-lesion errors.} 
  \label{fig2}
\end{figure*}

% Table~\ref{tab:qata_main} reports results on QaTa-COV19 and MosMedData+ under different label ratios. We compare Semi-MedRef with four groups of methods: (i) natural-image supervised RES (LAVT), (ii) natural-image semi-supervised RES (SemiRES), (iii) medical-image semi-supervised referring segmentation (LViT and TextMatch), and (iv) medical supervised backbones (GuideDecoder and MMI-UNet), on which we instantiate Semi-MedRef. Since TextMatch is not open-sourced, its 5\% and 15\% results are taken from the original paper and other ratios are left blank.
Table~\ref{tab:qata_main} compares Semi-MedRef with three groups
of methods: (i) natural-image RES methods (LAVT~\cite{yang2022lavt} and
SemiRES); (ii) supervised medical VLMs (ViTexNet, FMISeg, MG-UNet, GuideDecoder, and
MMI-UNet); and (iii) medical semi-supervised methods (LViT, TextMatch, DuSSS~\cite{Pan_Qiao_Lou_Ji_Li_2025}, AMVLM~\cite{pan2025amvlm}, TextMoE, and TeViD~\cite{zeng2026tevid}). %At each labeled ratio, all methods are reproduced using the same labeled split, and SSL methods additionally share the same complementary unlabeled set. 
For each labeled ratio, all baseline methods are reproduced using the same randomly sampled labeled subset, while SSL methods additionally use the complementary unlabeled subset.
For Semi-MedRef with 100\% labels, the unlabeled consistency branch is disabled, while the supervised augmentations and PACL remain active. We evaluate Semi-MedRef with two backbones, MMI-UNet and TextMoE-Seg, while the results on other backbones are reported in supplementary material.

Across both datasets, Semi-MedRef achieves SOTA Dice and mIoU at every evaluated label ratio. Its advantage over SSL baselines such as TeViD stems from synchronizing image, text, and pseudo-mask perturbations while using PACL to preserve partial positional compatibility, thereby reducing misaligned pseudo-supervision. Furthermore, Semi-MedRef is not tied to a specific segmentation architecture: it consistently improves MMI-UNet and TextMoE-Seg in Table~\ref{tab:qata_main}, with corresponding gains on additional backbones reported in the supplementary material. Qualitatively, Fig.~\ref{fig2} shows Semi-MedRef produces cleaner boundaries and finer lesion details while reducing background leakage.

% \begin{table}[t] 
% \centering 
% {\small \setlength{\tabcolsep}{1pt} \renewcommand{\arraystretch}{1.0} 
% \begin{tabular}{@{}lcccccccc@{}} 
% \toprule \multirow{2}{*}{Method} & \multirow{2}{*}{\makecell{T-S\\EMA}} & \multirow{2}{*}{\makecell{Pos-\\Aug}} & \multirow{2}{*}{\makecell{T-\\PatchMix}} & \multirow{2}{*}{PACL} & \multicolumn{2}{c}{1\%} & \multicolumn{2}{c}{2\%} \\ \cmidrule(lr){6-7} \cmidrule(lr){8-9} & & & & & Dice & mIoU & Dice  & mIoU \\ 
% \midrule Baseline & \ding{55} & \ding{55} & \ding{55} & \ding{55} & 81.02 & 68.09 & 84.63 & 73.36 \\ + T-S EMA & \ding{51} & \ding{55} & \ding{55} & \ding{55} & 84.58 & 73.27 & 85.75 & 75.06 \\ + PosAug & \ding{51} & \ding{51} & \ding{55} & \ding{55} & 85.89 & 75.27 & 86.59 & 76.35 \\ + T-PatchMix & \ding{51} & \ding{51} & \ding{51} & \ding{55} & 86.02 & 75.47 & 87.05 & 77.07 \\ \rowcolor{gray!20} + PACL (\textbf{Ours}) & \ding{51} & \ding{51} & \ding{51} & \ding{51} & \textbf{86.39} & \textbf{76.03} & \textbf{87.25} & \textbf{77.38} \\ 
% \bottomrule \end{tabular} } 
% \caption{Ablation studies on model components on the QaTa-COV19 dataset.} 
% \label{tab:ablation_lowratio} 
% \end{table}

\begin{table}[t] 
\centering 
{\small \setlength{\tabcolsep}{1pt} \renewcommand{\arraystretch}{1.0} 
\begin{tabular}{@{}lcccccccc@{}} 
\toprule \multirow{2}{*}{Method} & \multirow{2}{*}{\makecell{T-S\\EMA}} & \multirow{2}{*}{\makecell{T-\\PatchMix}} & \multirow{2}{*}{\makecell{Pos-\\Aug}} & \multirow{2}{*}{PACL} & \multicolumn{2}{c}{1\%} & \multicolumn{2}{c}{2\%} \\ \cmidrule(lr){6-7} \cmidrule(lr){8-9} & & & & & Dice & mIoU & Dice  & mIoU \\ 
\midrule Baseline & \ding{55} & \ding{55} & \ding{55} & \ding{55} & 81.02 & 68.09 & 84.63 & 73.36 \\ + T-S EMA & \ding{51} & \ding{55} & \ding{55} & \ding{55} & 84.58 & 73.27 & 85.75 & 75.06 \\ + T-PatchMix & \ding{51} & \ding{51} & \ding{55} & \ding{55} & 85.59 & 74.89 & 86.60 & 76.72 \\ + PosAug & \ding{51} & \ding{51} & \ding{51} & \ding{55} & 85.88 & 75.19 & 87.00 & 77.10 \\ \rowcolor{gray!20} + PACL (\textbf{Ours}) & \ding{51} & \ding{51} & \ding{51} & \ding{51} & \textbf{86.39} & \textbf{76.03} & \textbf{87.25} & \textbf{77.38} \\ 
\bottomrule \end{tabular} } 
\caption{Model component ablation on QaTa-COV19.} 
\label{tab:ablation_lowratio} 
\end{table}
% Across both datasets and all evaluated label ratios, Semi-MedRef consistently improves over its corresponding backbone and existing fully supervised and semi-supervised baselines. In particular, Semi-MedRef (MMI-UNet) achieves Dice scores of 87.25\% (2\%) and 91.39\% (100\%) on QaTa-COV19, and 74.51\% (5\%) and 78.70\% (100\%) on MosMedData+. Semi-MedRef also yields consistent gains when instantiated on GuideDecoder, indicating the improvements are not backbone-specific. Compared with medical semi-supervised baselines, Semi-MedRef outperforms LViT across all ratios and surpasses TextMatch under the reported settings (5\%, 15\%). Qualitatively, Fig.~\ref{fig2} shows that Semi-MedRef produces masks with cleaner boundaries and finer details with reduced leakage.
% \vspace{-2mm}

\subsection{Ablation Study}

% Table~\ref{tab:ablation_lowratio} analyzes the contribution of each component on QaTa-COV19 under 1\% and 2\% labeled settings. Starting from the supervised baseline, introducing the EMA teacher--student framework yields a clear improvement, validating the benefit of leveraging unlabeled data via consistency learning. Adding PosAug further boosts performance, suggesting improved robustness to positional language variations. Incorporating T-PatchMix provides additional gains, indicating that alignment-preserving patch mixing helps under strong perturbations. Finally, ITCL consistently brings the best results, confirming that position-guided contrastive alignment complements the segmentation consistency objective in low-label regimes. 
% Table~\ref{tab:ablation_aug_full} reports an augmentation ablation on QaTa-COV19 with 1\% labeled data. ImgStrong provides a clear benefit, as removing it reduces Dice from 84.58\% to 82.62\%. In contrast, applying vanilla CutMix degrades performance, consistent with the fact that mixed images can introduce text-inconsistent composites and thus weaken cross-modal alignment~\cite{zang2025resmatch}. The weak multimodal augmentation Semantic-Preserving Change (SPC) yields only a marginal improvement over the baseline. Our alignment-preserving augmentations PosAug and T-PatchMix are both consistently effective.
Table~\ref{tab:ablation_lowratio} reports component-wise ablations on QaTa-COV19, indicating each component contributes positive gains, with the full model performing best.
% Table~\ref{tab:ablation_aug_full} shows that our alignment-preserving multimodal augmentations (PosAug and T-PatchMix) outperform strong text-only augmentation, image-only CutMix, and the SPC baseline under 1\% labels, highlighting the benefit of preserving image–text alignment during strong perturbations. Moreover, using ITCL yields higher Dice than the CLIP-style loss, indicating that region-aware soft positives provide a more effective cross-modal alignment signal.
Table~\ref{tab:ablation_aug_full} summarizes augmentation and alignment ablations. Ours-V1 vs TextStrong (86.39\% vs. 85.90\%) shows that PosAug provides a more effective MRIS-specific text perturbation than generic text augmentations. Ours-V2 vs. CutMix (86.60\% vs. 84.75\%) demonstrates that T-PatchMix preserves image–text coherence, while naive CutMix disrupts alignment: its performance even drops below the T-S baseline. Finally, Ours-V3 vs. Ours-All (86.51\% vs. 87.25\%) confirms that PACL offers stronger cross-modal alignment than a CLIP-style loss.

Table~\ref{tab:tpatchmix_comparison} shows that position constraints improve spatial consistency, whereas probability guidance increases lesion score. Their combination achieves the best Dice, confirming that effective multimodal mixing benefits from both positional alignment and lesion-relevant
content.
\begin{table}[t]
\centering

{\small
\setlength{\tabcolsep}{1.0pt}
\renewcommand{\arraystretch}{1.0}

\begin{tabular}{@{}lccccccc@{}}
\toprule
Method
& \makecell{Img-\\Strong}
& \makecell{Text-\\Strong}
& PosAug
& \makecell{T-\\PatchMix}
& CLIP
& PACL
& \makecell{Dice $\uparrow$} \\
\midrule

T-S EMA
& \ding{51}
& \ding{55}
& \ding{55}
& \ding{55}
& \ding{55}
& \ding{55}
& 85.75 \\

TextStrong
& \ding{51}
& \ding{51}
& \ding{55}
& \ding{55}
& \ding{55}
& \ding{55}
& 85.90 \\

\rowcolor{gray!20}
w/ CutMix
& \ding{51}
& \ding{55}
& \ding{55}
& \ding{55}
& \ding{55}
& \ding{55}
& 84.75 \\

w/ SPC
& \ding{51}
& \ding{55}
& \ding{55}
& \ding{55}
& \ding{55}
& \ding{55}
& 85.93 \\

\midrule

Ours-V1
& \ding{51}
& \ding{55}
& \ding{51}
& \ding{55}
& \ding{55}
& \ding{55}
& 86.39 \\

\rowcolor{gray!20}
Ours-V2
& \ding{51}
& \ding{55}
& \ding{55}
& \ding{51}
& \ding{55}
& \ding{55}
& 86.60 \\

\rowcolor{gray!50}
Ours-V3
& \ding{51}
& \ding{55}
& \ding{51}
& \ding{51}
& \ding{51}
& \ding{55}
& 86.51 \\

\rowcolor{gray!50}
Ours-All
& \ding{51}
& \ding{55}
& \ding{51}
& \ding{51}
& \ding{55}
& \ding{51}
& \textbf{87.25} \\

\bottomrule
\end{tabular}
}

\caption{Augmentation and alignment ablations on QaTa-COV19 with 2\% labels. Img-Strong includes
ColorJitter and GaussianBlur. Text-Strong includes Synonym Replacement
and Random Deletion. SPC follows RESMatch~\cite{zang2025resmatch} by
editing positional words to match image transformations.}
\label{tab:ablation_aug_full}
\end{table}

\begin{figure}[t]
    \centering
    \includegraphics[width=\columnwidth]{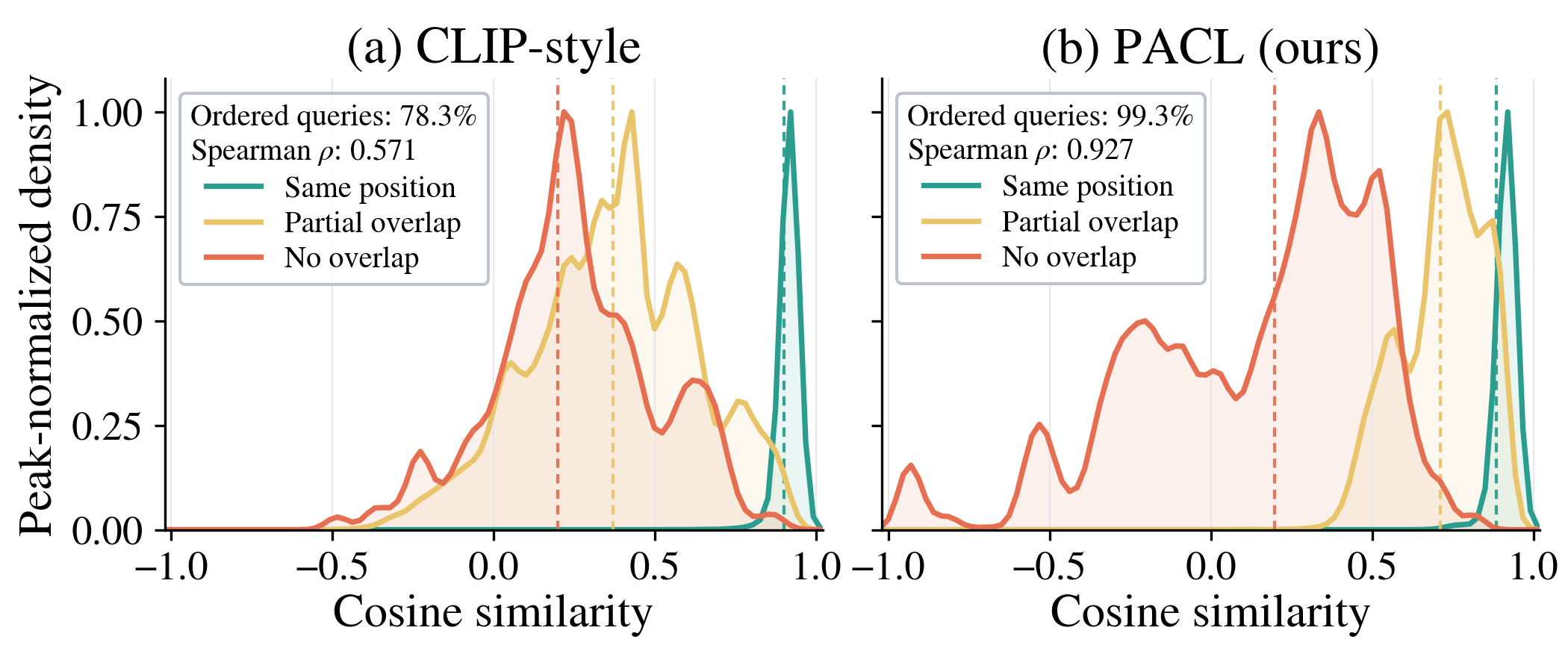}
    \caption{Cosine-similarity distributions of QaTa-COV19 validation pairs grouped by positional overlap (same, partial, or none) under CLIP-style alignment and PACL. PACL better separates partial-overlap from no-overlap pairs. For an ordered query, mean similarity decreases from same-position to partial-overlap to no-overlap texts.}
    \label{fig:itcl_similarity}
\end{figure}

\begin{table}[t]
\centering
{\small
\setlength{\tabcolsep}{2.0pt}
\renewcommand{\arraystretch}{1.0}
\begin{tabular}{@{}lccc@{}}
\toprule
Mixing strategy
& Dice $\uparrow$
& Lesion score $\uparrow$
& Position overlap $\uparrow$ \\
\midrule
Random PatchMix
& 85.01
& 17.98
& 34.45 \\

T-PatchMix-pos
& 86.96
& 19.81
& \textbf{100.00} \\

T-PatchMix-prob
& 85.98
& 68.10
& 41.77 \\

T-PatchMix-hybrid
& \textbf{87.25}
& \textbf{79.24}
& 99.99 \\
\bottomrule
\end{tabular}
}

\caption{Patch-mixing comparison (\%) on QaTa-COV19 with 2\% labels. Lesion score is mean teacher-predicted foreground probability within each patch; position overlap is fraction of patch pixels inside the coarse region specified by $T_j^u$.}
\label{tab:tpatchmix_comparison}
\end{table}

\paragraph{Cross-modal alignment analysis.}
Figure~\ref{fig:itcl_similarity} compares cosine-similarity distributions of image–text pairs grouped by positional overlap under CLIP and PACL. As can be seen, PACL organizes cross-modal similarity according to positional overlap. Under PACL, the partial-overlap distribution (yellow) lies close to the same-position distribution (green; mean gap 0.176) and is clearly separated from the no-overlap distribution (red; mean gap 0.524). In contrast, under the CLIP-style objective, the partial-overlap distribution is far from the same-position one (mean gap 0.480) but close to the no-overlap one (mean gap 0.154). Accordingly, PACL achieves 99.3\% correctly ordered queries and a Spearman correlation of 0.927, compared with 78.3\% and 0.571 for CLIP, showing that PACL preserves partially shared anatomical semantics rather than treating all unmatched referring texts as equally negative.

% Figure~\ref{fig:itcl_similarity} examines whether the learned
% cross-modal similarity reflects graded anatomical compatibility.
% Under the CLIP-style objective, the mean similarity of
% partial-overlap pairs is much closer to that of no-overlap pairs
% than to that of same-position pairs, with corresponding gaps of
% 0.154 and 0.480. This suggests that partially compatible reports
% are still represented largely as negatives. ITCL reverses this
% geometry: the partial-overlap distribution lies only 0.176 from
% the same-position distribution but 0.524 from the no-overlap
% distribution. Accordingly, the fraction of queries satisfying
% $s_{\mathrm{same}}>s_{\mathrm{partial}}>s_{\mathrm{none}}$
% increases from 78.3\% to 99.3\%, while the Spearman correlation
% between cosine similarity and positional overlap increases from
% 0.571 to 0.927. These results demonstrate that ITCL preserves
% partially shared anatomical semantics and organizes the embedding
% space according to graded positional compatibility, rather than
% treating all unmatched reports as equally negative.

\paragraph{Hyperparameter Sensitivity Analysis.}
Table~\ref{tab:hyperparameter_sensitivity} shows varying either PACL weight changes Dice at most 0.28 percentage points, indicating low sensitivity over the evaluated ranges. Analyses for other hyperparameters (e.g., PACL temperature $\tau$, pseudo-label threshold $\delta$ and soft-threshold strategy) are reported in the supplementary material.

% As shown in Table~\ref{tab:hyperparameter_sensitivity}, Semi-MedRef remains stable across reasonable variations in the PACL temperature and the loss weights for labeled and unlabeled samples. The Dice score deviates from the default configuration by at most 0.38 percentage points across all evaluated settings, with some alternatives yielding marginal improvements. Moreover, soft pseudo-label formulations do not consistently outperform hard thresholding, demonstrating that the proposed method is robust to hyperparameter variations and is not highly sensitive within the evaluated ranges.
\begin{table}[t]
    \centering
    {\small
        \setlength{\tabcolsep}{0.8mm}
        \begin{tabular}{@{}llccc@{}}
            \toprule
            Hyperparameter
            & Setting
            & Dice (\%) $\uparrow$
            & mIoU (\%) $\uparrow$
            & $\Delta$ \\
            \midrule

            \multirow{3}{*}{
                \makecell[l]{Supervised PACL \\weight
                $\lambda_{\mathit{pacl}}^{\mathit{sup}}$}
            }
                & 0.01
                & 87.04
                & 77.05
                & $-0.21$ \\
                & $0.02^{\dagger}$
                & \textbf{87.25}
                & \textbf{77.38}
                & --- \\
                & 0.04
                & 86.97
                & 76.92
                & $-0.28$ \\

            \midrule

            \multirow{3}{*}{
                \makecell[l]{Unsupervised PACL \\weight
                $\lambda_{\mathit{pacl}}^{\mathit{unsup}}$}
            }   
                & 0.05
                & 87.14
                & 77.21
                & $-0.11$ \\
                & $0.10^{\dagger}$
                & \textbf{87.25}
                & \textbf{77.38}
                & --- \\
                & 0.20
                & 87.12
                & 77.18
                & $-0.13$ \\

            \bottomrule
        \end{tabular}
    }
    \caption{Hyperparameter sensitivity analysis on QaTa-COV19 with 2\% labels. $\dagger$ denotes the default; bold denotes the best result, and $\Delta$ is the Dice change from default.}
    \label{tab:hyperparameter_sensitivity}
\end{table}

\paragraph{Importance of Positional Words.}
Controlled validation-set interventions quantify the positional-language sensitivity observed in Fig.~\ref{fig:motivation}. Removing positional phrases substantially decreases Dice, whereas masking the same number of non-positional tokens causes no degradation; left--right swaps also induce directionally consistent localization changes in unilateral cases. These results confirm the importance of positional language for spatial grounding and support the designs of PosAug, T-PatchMix, and PACL, with further analyses provided in the supplementary material.

\subsection{Generality Across Medical Domains}
\begin{figure}[!t]
  \centering
  \includegraphics[width=0.85\columnwidth]{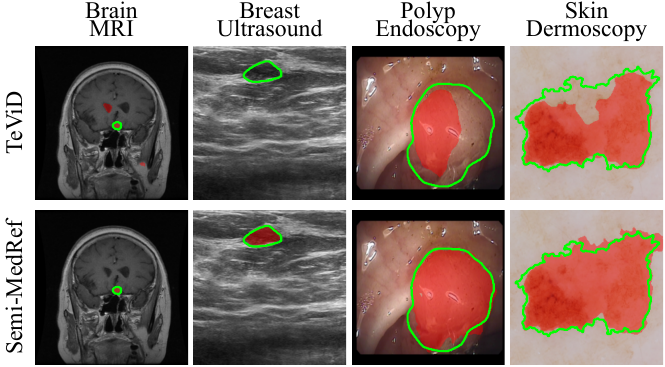}
  \caption{Qualitative comparisons on PosMed. Red masks denote predictions and green contours denote ground truth.
  %Predictions are shown as red translucent masks, while ground-truth annotations are indicated by green contours.
} 
  \label{posmed}
\end{figure}

To assess the generality of Semi-MedRef beyond chest-lesion benchmarks,
we evaluate it on the public dataset PosMed~\cite{trinh2026prsmed}, which spans heterogeneous anatomies and modalities, including brain MRI, breast ultrasound, polyp endoscopy, and skin dermoscopy, using PosMed's native referring expressions with explicit spatial cues. More details about the dataset are provided in the supplementary material. For each comparing method, a single model is jointly trained across all four domains using the same 15\% labeled split; SSL methods additionally use the remaining 85\% training data as unlabeled samples. As shown in Table~\ref{tab:posmed_cross_domain}, Semi-MedRef achieves the best Dice and mIoU across domains, outperforming both MMI-UNet and TeViD. These consistent improvements suggest that Semi-MedRef generalizes to heterogeneous anatomies, imaging modalities, and referring expressions, rather than relying on the structured chest-report style of previous benchmarks. Fig.~\ref{posmed} further shows that Semi-MedRef produces more complete lesion coverage and tighter boundaries.

% \begin{table}[t]
%     \centering
%     {\footnotesize
%     \setlength{\tabcolsep}{1.5pt}
%     \begin{tabular}{@{}lccccc@{}}
%         \toprule
%         Method
%         & \makecell{Brain\\MRI}
%         & \makecell{Breast\\US}
%         & \makecell{Polyp\\Endoscopy}
%         & \makecell{Skin\\Dermoscopy}
%         \\
%         \midrule
%         MMI-UNet  & 65.25/48.43 & 82.37/70.03 & 76.31/61.70 & 91.56/84.43 \\
%         TextMoE & 63.60/46.62 & 82.48/71.12 & 75.99/61.28 & 90.60/84.22 \\
%         Semi-MedRef & \textbf{67.00/50.37} & \textbf{84.90/73.76} & \textbf{77.54/63.31} & \textbf{91.92/85.05} \\
%         \bottomrule
%     \end{tabular}}
%     \caption{Cross-domain performance on the four non-lung lesion-centric PosMed subsets using 15\% labeled data. Each entry reports Dice/mIoU (\%).}
%     \label{tab:posmed_cross_domain}
% \end{table}
\begin{table}[t]
    \centering
    {\footnotesize
    \setlength{\tabcolsep}{1.5pt}
    \begin{tabular}{@{}lccccc@{}}
        \toprule
        Method
        & \makecell{Brain\\MRI}
        & \makecell{Breast\\US}
        & \makecell{Polyp\\Endoscopy}
        & \makecell{Skin\\Dermoscopy}
        \\
        \midrule
        MMI-UNet  & 65.25/48.43 & 82.37/70.03 & 76.31/61.70 & 91.56/84.43 \\
        TeViD & 62.26/45.20 & 70.03/53.88 & 72.47/56.83 & 87.93/78.46 \\
        Semi-MedRef & \textbf{67.00/50.37} & \textbf{84.90/73.76} & \textbf{77.54/63.31} & \textbf{91.92/85.05} \\
        \bottomrule
    \end{tabular}}
    \caption{Multi-domain performance on the four non-lung lesion-centric PosMed subsets using 15\% labeled data. Each entry reports Dice/mIoU (\%).}
    \label{tab:posmed_cross_domain}
\end{table}
\begin{table}[t!]
\centering

{\small
\setlength{\tabcolsep}{2.0pt}
\renewcommand{\arraystretch}{1.05}

\begin{tabular}{@{}lcccc@{}}
\toprule
Labels
& Standard
& Zero
& ISPG
& Oracle \\
\midrule

2\%
& 87.25/77.38
& 81.82/69.23
& 86.12/75.62
& 87.01/77.01 \\

5\%
& 88.31/78.98
& 75.80/61.03
& 87.55/77.85
& 88.45/79.29 \\

15\%
& 89.84/81.56
& 74.86/59.82
& 88.57/79.48
& 89.75/81.40 \\

\midrule
\multicolumn{2}{@{}l}{Prediction-head overhead}
& \multicolumn{3}{c@{}}
{$+0.30$M Params \quad /\quad$<0.001$ GFLOPs} \\

\bottomrule
\end{tabular}
}

\caption{Robustness to removing positional expressions on QaTa-COV19
(Dice/mIoU, \%). Standard uses complete referring texts; Zero, ISPG,
and Oracle use position-stripped texts with zero, image-predicted,
and text-derived position vectors, respectively. The last row shows extra inference cost.}
\label{tab:ispg_robustness}
\end{table}
\subsection{Robustness to Missing Positional Expressions}

We evaluate ISPG after removing all explicit positional
expressions while retaining the remaining referring text.
The GAP--MLP predictor and a 50\% training removal rate are
selected on the validation set; additional analysis is provided in the supplementary material. As shown in Table~\ref{tab:ispg_robustness}, ISPG consistently narrows the performance gap between the Zero and Oracle settings, with particularly clear benefits under limited supervision. Its performance remains close to the text-derived Oracle,
demonstrating that visually recoverable anatomical cues can
effectively compensate for missing positional language. The
additional inference cost is incurred only when this optional
robustness branch is enabled.

\FloatBarrier
\section{Conclusion}
We presented Semi-MedRef, a semi-supervised MRIS framework that preserves image--text alignment under strong perturbations. It combines alignment-aware PosAug and T-PatchMix with PACL's graded
positional supervision, while optional ISPG improves robustness to missing positional expressions. Experiments on two MRIS datasets, multiple backbones, and heterogeneous PosMed domains show consistent gains across label regimes and medical domains.

\bibliography{aaai2027}

\clearpage
\twocolumn[
\begin{center}
{\LARGE\bfseries Supplementary Material\par}
\vspace{1.5em}
\end{center}
]
\appendix

\urlstyle{rm} % DO NOT CHANGE THIS
\def\UrlFont{\rm} % DO NOT CHANGE THIS

\frenchspacing % DO NOT CHANGE THIS

\pdfinfo{
/TemplateVersion (2027.1)
}

\setcounter{secnumdepth}{2}

\title{Supplementary Material for\\
Semi-MedRef: Semi-Supervised Medical Referring Image Segmentation with
Cross-Modal Alignment}
\author{Anonymous Submission}
\affiliations{}

\appendix

\section{Experimental and Reproducibility Details}
\label{app:reproducibility}

\subsection{Computing Infrastructure}

All experiments were executed on one server equipped with four NVIDIA
GeForce RTX 3090 GPUs, each with 24~GiB of memory. Each individual training or
evaluation run used exactly one GPU; the four devices were used only to run
independent experiments in parallel and were not combined through
data-parallel training. The host contained two Intel Xeon Silver 4314
processors (16 physical cores per socket and 64 logical CPUs in total),
125~GiB of RAM, and ran Ubuntu 20.04.6 LTS with Linux kernel
5.4.0-216-generic. The NVIDIA driver version was 535.247.01.

Table~\ref{tab:software-environment} records the software environment used to
run the experiments. The released \texttt{environment.yml} and
\texttt{requirements.txt} files provide installable specifications of this
environment.

\begin{table*}[t!]
\centering
\small
\setlength{\tabcolsep}{5pt}
\begin{tabular}{llll}
\toprule
Component & Specification & Component & Version \\
\midrule
GPU server & $4\times$ RTX 3090, 24~GiB each
 & Python & 3.10.15 \\
Per-run allocation & One RTX 3090
 & PyTorch / TorchVision & 2.5.0+cu121 / 0.20.0+cu121 \\
CPU & $2\times$ Xeon Silver 4314
 & CUDA runtime / cuDNN & 12.1 / 9.1.0 \\
Host memory & 125~GiB
 & PyTorch Lightning & 1.9.0 \\
Operating system & Ubuntu 20.04.6 LTS
 & TorchMetrics / MONAI & 0.10.3 / 1.0.1 \\
Linux kernel & 5.4.0-216-generic
 & Transformers & 4.30.2 \\
NVIDIA driver & 535.247.01
 & NumPy / pandas & 1.26.4 / 2.3.0 \\
 &  & einops / scikit-learn & 0.8.1 / 1.7.0 \\
\bottomrule
\end{tabular}
\caption{Reference hardware and software environment. Each model run used one
RTX 3090; up to four independent runs were executed concurrently.}
\label{tab:software-environment}
\end{table*}

\subsection{Datasets and Splits}

We used the public QaTa-COV19 and MosMedData+ splits described in the main
paper; Table~\ref{tab:data-splits} gives the exact image--report counts. Only
the training partition was divided into mask-labeled and mask-unlabeled pools.
Validation selected checkpoints, while the test partition was used only for
final evaluation and post-hoc statistics; neither entered either training pool.

\begin{table*}[t!]
\centering
\small
\setlength{\tabcolsep}{6pt}
\begin{tabular}{lrrrrrrr}
\toprule
\multirow{2}{*}{Dataset}
& \multirow{2}{*}{Train}
& \multirow{2}{*}{Validation}
& \multirow{2}{*}{Test}
& \multicolumn{4}{c}{Mask-labeled training samples} \\
\cmidrule(lr){5-8}
& & & & 2\% & 5\% & 15\% & 100\% \\
\midrule
QaTa-COV19 & 5,716 & 1,429 & 2,113 & 114 & 286 & 857 & 5,716 \\
MosMedData+ & 2,183 & 273 & 273 & 43 & 109 & 327 & 2,183 \\
\bottomrule
\end{tabular}
\caption{Dataset split sizes and labeled-training-set sizes at each mask
annotation ratio. Train, Validation, and Test report the total numbers of
samples in the corresponding partitions. The 2\%, 5\%, 15\%, and 100\%
columns report the numbers of mask-labeled training samples; the remaining
training samples form the mask-unlabeled pool.}
\label{tab:data-splits}
\end{table*}

The released low-label manifests were generated once with split seed 42 and
fixed for all model seeds. Grouping by image identifier prevented reports of
one image from crossing the labeled/unlabeled boundary.

The six-dimensional multi-hot targets are ordered as left-upper, left-middle, left-lower, right-upper, right-middle, and right-lower. For MosMedData+, reports were lowercased and whitespace-normalized; a target entry was set to 1 when its side and vertical term occurred in a location phrase (e.g., “upper left lung”), and otherwise to 0. No manual labels were added. The image-derived predictor used
only training-report targets; no validation or test report label was optimized.

\subsection{Input Processing and Augmentation}

Images and masks were resized to $224\times224$, and reports were tokenized
with CXR-BERT to 24 tokens. Training used joint RandomZoom
($[0.95,1.20]$, $p=0.1$), ColorJitter (brightness and contrast 0.2,
$p=0.8$), and Gaussian blur (kernel 7, $\sigma\in[0.1,2.0]$, $p=0.5$).
Horizontal flipping was disabled to preserve anatomical laterality.
PosAug and T-PatchMix settings are given in
Table~\ref{tab:final-hyperparameters}. No augmentation or test-time
augmentation was used during validation or testing.

\subsection{Hyperparameters and Optimization}

\paragraph{Final configuration.}
Table~\ref{tab:final-hyperparameters} summarizes the final hyperparameters
used for the main Semi-MedRef experiments.
\begin{table*}[t!]
\centering
\small
\setlength{\tabcolsep}{5pt}
\begin{tabular}{lll}
\toprule
Category & Hyperparameter & Final value \\
\midrule
Architecture
 & Image/text encoders & ConvNeXt-Tiny / CXR-BERT \\
 & Contrastive projection dimension & 768 \\
\midrule
Optimization
 & Effective training batch / validation batch & 32 / 8 \\
 & Precision & FP32 \\
 & Optimizer and initial learning rate & AdamW, $3\times10^{-4}$ \\
 & Weight decay and Adam coefficients & $10^{-2}$; $(0.9,0.999)$ \\
 & Learning-rate schedule & cosine, $T_{\max}=200$, minimum $10^{-6}$ \\
 & Epoch range and early stopping & 20--100; patience 30, min.\ change $10^{-3}$ \\
\midrule
Teacher--student
 & Burn-in / unsupervised ramp-up & 5 / 15 epochs \\
 & Maximum unlabeled segmentation weight & 0.3 \\
 & Hard pseudo-mask threshold & 0.55 \\
 & EMA momentum & $0.99\rightarrow0.999$ over 20 epochs; final 0.999 \\
 & Labeled/unlabeled loader combination & maximum-size cycling \\
\midrule
Cross-modal training
 & PosAug probability & 0.5 \\
 & T-PatchMix block & $16\times16$ \\
 & PACL temperature $\tau$ & 0.07 \\
 & Supervised PACL weight $\lambda_{\mathit{pacl}}^{\mathit{sup}}$ & 0.02 \\
 & Unsupervised PACL weight $\lambda_{\mathit{pacl}}^{\mathit{unsup}}$ & 0.10 \\
\bottomrule
\end{tabular}
\caption{Final hyperparameters for the main Semi-MedRef experiments. The same
settings were used for both datasets and both segmentation backbones unless
explicitly stated.}
\label{tab:final-hyperparameters}
\end{table*}

\paragraph{Optimization protocol.}
Both supervised and unsupervised segmentation used MONAI
Dice--cross-entropy loss. During the five-epoch burn-in, the unlabeled
segmentation weight was zero and then increased linearly to 0.3 over
15 epochs. The EMA momentum increased from 0.99 to 0.999 over 20 epochs.
The checkpoint with the highest validation Dice was evaluated. At 100\%
labels, unlabeled segmentation, unsupervised PACL, and T-PatchMix were
disabled, while labeled segmentation and supervised PACL were retained.

\paragraph{Sensitivity analysis.}
Complementing the 2\%-label PACL-weight analysis in Table~5 of the main
paper, we additionally evaluated PACL-related choices on QaTa-COV19 with
5\% labels in Table~\ref{tab:hyperparameter-sensitivity}. Hyperparameters were selected on validation set. 

For the pseudo-label construction comparison, let $p_i(x)$ denote the
teacher foreground probability at pixel $x$. The hard and soft targets are
defined as:
\begin{equation}
\begin{aligned}
\widetilde{y}_i^{\mathrm{hard}}(x)
&= \mathbb{I}\!\left[p_i(x) \geq \delta\right], \\
\widetilde{y}_{i,T}^{\mathrm{soft}}(x)
&= \sigma\!\left(\frac{p_i(x)-\delta}{T}\right),
\end{aligned}
\label{eq:hard_soft_pseudo_labels}
\end{equation}
where $T$ controls the transition sharpness around $\delta$. Both targets
are optimized with the same unsupervised segmentation loss. 

The maximum Dice difference from the
default within each factor is 0.38 percentage points, and soft
pseudo-labeling provides no consistent improvement over hard thresholding.
We further varied the hard pseudo-mask confidence threshold $\delta$ under
the finalized protocol. As shown in
Table~\ref{tab:delta-sensitivity}, performance remains stable over a broad
range: relative to the default $\delta=0.55$, Dice
changes by only $-0.20$ to $+0.15$ percentage points.

\begin{table}[t!]
\centering
{\small
\setlength{\tabcolsep}{1mm}
\begin{tabular}{@{}llccc@{}}
\toprule
Factor & Setting & Dice & mIoU & $\Delta$ \\
\midrule
PACL temp. $\tau$
 & 0.03 & 88.01 & 78.58 & $-0.30$ \\
 & 0.05 & 88.17 & 78.85 & $-0.14$ \\
 & $0.07^{\dagger}$ & \textbf{88.31} & \textbf{78.98} & --- \\
\midrule
\multirow{2}{*}{
\makecell[l]{Supervised PACL\\weight
$\lambda_{\mathit{pacl}}^{\mathit{sup}}$}}
 & $0.02^{\dagger}$ & \textbf{88.31} & \textbf{78.98} & --- \\
 & 0.04 & 88.01 & 78.59 & $-0.30$ \\
\midrule
\multirow{3}{*}{
\makecell[l]{Unsupervised PACL\\weight
$\lambda_{\mathit{pacl}}^{\mathit{unsup}}$}}
 & $0.10^{\dagger}$ & 88.31 & 78.98 & --- \\
 & 0.20 & \textbf{88.36} & \textbf{79.15} & $+0.05$ \\
 & 0.40 & 88.18 & 78.86 & $-0.13$ \\
\midrule
\multirow{3}{*}{\makecell[l]{Pseudo-label\\thresholding}}
 & Soft ($T=.05$) & 87.93 & 78.46 & $-0.38$ \\
 & Soft ($T=.10$) & 88.09 & 78.71 & $-0.22$ \\
 & Hard$^{\dagger}$ & \textbf{88.31} & \textbf{78.98} & --- \\
\bottomrule
\end{tabular}
}
\caption{Additional PACL hyperparameter and pseudo-label-thresholding
sensitivity on QaTa-COV19 with 5\% labels. $\dagger$ marks the default; bold
indicates the best result within each factor. $\Delta$ is the Dice change
from default in percentage points.
$\lambda_{\mathit{pacl}}^{\mathit{sup}}$ and
$\lambda_{\mathit{pacl}}^{\mathit{unsup}}$ weight the supervised and
unsupervised PACL losses, respectively.}
\label{tab:hyperparameter-sensitivity}
\end{table}

\begin{table}[t!]
\centering
{\small
\setlength{\tabcolsep}{4.5mm}
\begin{tabular}{@{}cccc@{}}
\toprule
$\delta$ & Dice & mIoU & $\Delta$Dice \\
\midrule
0.45 & 88.46 & 79.31 & $+0.15$ \\
0.50 & 88.27 & 79.01 & $-0.04$ \\
$0.55^{\dagger}$ & 88.31 & 78.98 & --- \\
0.60 & 88.11 & 78.74 & $-0.20$ \\
0.65 & 88.17 & 78.84 & $-0.14$ \\
0.75 & 88.32 & 79.08 & $+0.01$ \\
\bottomrule
\end{tabular}
}
\caption{Sensitivity to the hard pseudo-mask confidence threshold $\delta$
on QaTa-COV19 with 5\% labels under the finalized protocol.
$\Delta$Dice is measured relative to the default
$\delta=0.55$; $\dagger$ marks this default.}
\label{tab:delta-sensitivity}
\end{table}

\subsection{Randomness, Number of Runs, and Model Selection}
For both Semi-MedRef rows in main-paper Table~1, the adjacent ``$\pm$ Std.''
row reports the sample standard deviation over model seeds 0, 1, and 42. All model seeds used the same seed-42 labeled manifests; changing
the model seed affected initialization, mini-batch order, stochastic
augmentation, and data-loader workers, but not the labeled cases.

Lightning seeded the main process and data-loader workers. Deterministic
cuDNN execution was enabled and cuDNN benchmarking was disabled. Checkpoints
were selected by maximum validation Dice, with early stopping after 30
validation epochs without an improvement of at least $10^{-3}$. Final
evaluation used the fixed test manifest and student weights.

To provide the additional-backbone evaluation promised in the main text, we
instantiate Semi-MedRef with GuideDecoder, ViTexNet, FMISeg, MG-UNet, and
TeViD. Table~\ref{tab:additional-backbones} compares each transferred model
with its native backbone under the same data splits and label budgets.

\begin{table*}[t!]
\centering
{\footnotesize
\setlength{\tabcolsep}{1.8pt}
\renewcommand{\arraystretch}{1.02}
\begin{tabular}{@{}l*{8}{c}@{}}
\toprule
\multirow{2}{*}{Method}
& \multicolumn{4}{c}{\textbf{QaTa-COV19}}
& \multicolumn{4}{c}{\textbf{MosMedData+}} \\
\cmidrule(lr){2-5}\cmidrule(lr){6-9}
& 2\% & 5\% & 15\% & 100\% & 2\% & 5\% & 15\% & 100\% \\
\midrule
GuideDecoder
& 80.44/67.28 & 84.23/72.76 & 85.03/73.95 & 89.78/81.45
& 66.34/49.63 & 71.21/55.30 & 73.10/57.95 & 77.75/63.60 \\
\quad + Semi-MedRef
& \textbf{86.07/75.54} & \textbf{87.12/77.18}& \textbf{89.13/80.39} & \textbf{90.91/83.34}
& \textbf{72.87/57.32} & \textbf{73.11/57.62} & \textbf{74.70/59.61} & \textbf{77.94/63.86} \\
\midrule
ViTexNet
& 62.08/45.01 & 75.42/60.54 & 88.29/79.04 & 89.71/81.35
& 62.34/45.28 & 67.90/52.13 & 71.11/55.62 & 77.39/63.46 \\
\quad + Semi-MedRef
& \textbf{86.34/75.97} & \textbf{87.94/78.47} & \textbf{88.66/79.63} & \textbf{90.25/82.24}
& \textbf{69.66/53.44} &\textbf{71.94/56.17} &\textbf{73.88/58.58} &\textbf{78.24/64.26} \\
\midrule
FMISeg
& 83.99/72.40 & 86.49/76.17 & 88.07/78.68 & 90.82/83.19
& 63.49/46.51 & 65.65/48.87 & 73.68/58.35& 76.30/61.44 \\
\quad + Semi-MedRef
& \textbf{85.77/75.04} & \textbf{87.49/77.76} & \textbf{89.37/80.77} & \textbf{91.18/83.79}
& \textbf{70.12/53.99} & \textbf{70.60/54.56} & \textbf{74.46/59.22} & \textbf{77.29/62.88} \\
\midrule
MG-UNet
& 83.13/71.13 & 86.48/76.19 & 88.02/78.60 & 90.76/83.09
& 59.53/42.38 & 67.81/51.30 & 71.37/55.48 & 76.39/61.79 \\
\quad + Semi-MedRef
& \textbf{87.03/77.03} & \textbf{87.94/78.48} & \textbf{89.05/80.26} & \textbf{90.77/83.11}
& \textbf{72.52/56.89} & \textbf{73.07/57.57} & \textbf{74.20/58.98} & \textbf{77.64/63.45} \\
\midrule
TeViD
& 84.50/73.15 & 86.10/75.59 & 88.15/78.82 & 90.51/82.69
& 66.99/50.37 & 72.15/56.43 & 75.18/60.23 & 77.60/63.40 \\
\quad + Semi-MedRef
& \textbf{87.61/77.97} & \textbf{88.14/78.80} & \textbf{89.76/81.41} & \textbf{91.31/84.02}
& \textbf{71.72/55.94} & \textbf{74.77/59.72} & \textbf{76.30/61.68} & \textbf{78.13/64.13} \\
\bottomrule
\end{tabular}}
\caption{Additional backbone-transfer results under the protocol of Table~1
in the main paper. Each entry reports mean Dice/mIoU (\%) over three model seeds. Bold marks the better
result within each native-backbone/Semi-MedRef pair. ISPG is disabled.}
\label{tab:additional-backbones}
\end{table*}

Semi-MedRef improves Dice of all backbone--dataset--label-ratio
comparisons. This pattern shows
that the largest and most consistent transfer gains arise in the low-label regime.

\paragraph{TextMoE-Seg Transfer Configuration.}
\label{app:textmoe-config}
The Ours (TextMoE-Seg) row in main-paper Table~1 is a backbone-transfer
experiment. We retain TextMoE's report-conditioned mixture-of-experts
architecture and replace its native semi-supervised training with
Semi-MedRef. It uses the same fixed data manifests, validation-based
checkpoint selection, input resolution, optimizer, and training schedule as
the MMI-UNet experiments, while retaining TextMoE-specific architectural
settings. ISPG is disabled in these comparisons. The TeViD transfer results follow the same backbone-transfer principle.

\paragraph{Why DuSSS and AMVLM are not treated as backbones.}
DuSSS and AMVLM are complete SSL pipelines rather than interchangeable
report-conditioned segmentation backbones. Their pretrained VLMs generate
text-derived supervision during training, whereas inference uses an
image-only U-Net. Replacing only the U-Net would not preserve the original
method, while retaining the complete pipeline would confound backbone
transfer with additional pretraining and auxiliary supervision. We
therefore retain them as native SSL baselines in the main paper.

\subsection{Additional Qualitative Results}
\label{app:additional-qualitative}

\begin{figure*}[t!]
\centering
\includegraphics[width=\textwidth]
{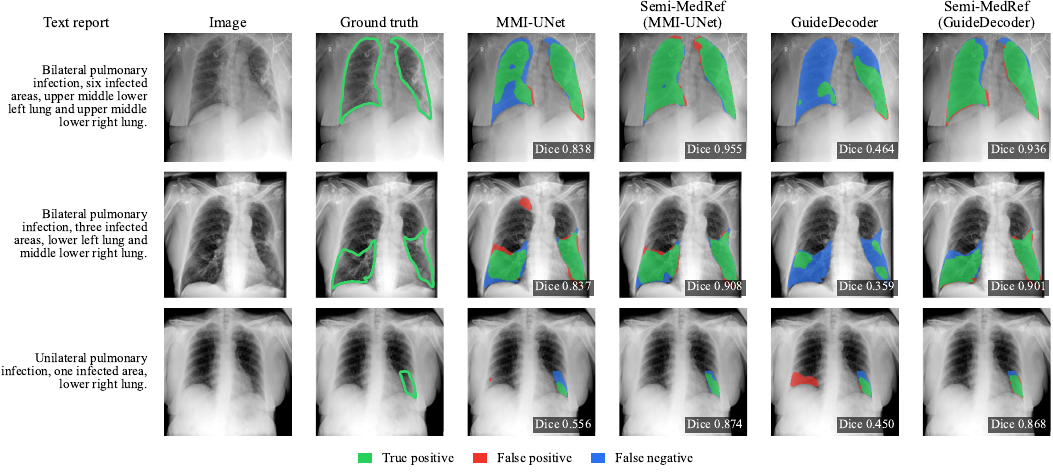}
\caption{Additional qualitative comparisons on QaTa-COV19 test
set under the 2\% labeled setting. The first column shows the corresponding
referring report. For each of the two segmentation backbones, we compare its
native baseline with the corresponding Semi-MedRef model. Green, red, and
blue indicate true-positive, false-positive, and false-negative regions,
respectively. Dice denotes the per-image overlap score.}
\label{fig:qualitative-qata}
\end{figure*}

\begin{figure*}[t!]
\centering
\includegraphics[width=\textwidth]
{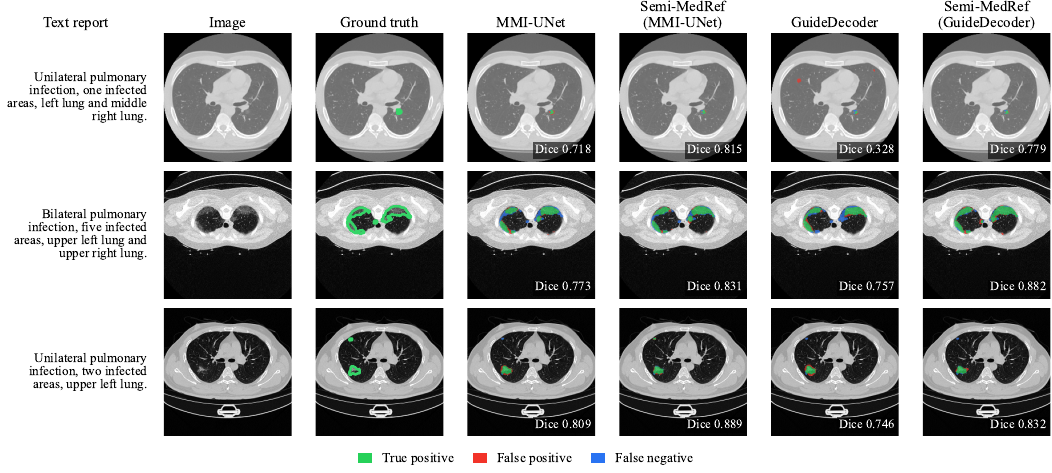}
\caption{Additional qualitative comparisons on MosMedData+ test
set under the 2\% labeled setting. The organization and color encoding follow
Figure~\ref{fig:qualitative-qata}. Across both MMI-UNet and GuideDecoder,
Semi-MedRef produces predictions with higher per-image Dice and closer
ground-truth alignment.}
\label{fig:qualitative-mosmed}
\end{figure*}

Figures~\ref{fig:qualitative-qata} and
\ref{fig:qualitative-mosmed} provide additional qualitative comparisons
under the 2\% labeled setting on the test sets of QaTa-COV19 and
MosMedData+, respectively. The corresponding referring report is shown in
the first column. Green, red, and blue denote true-positive, false-positive,
and false-negative regions, respectively.

Across both datasets and segmentation backbones, Semi-MedRef recovers more
of the target regions and reduces missed or spurious predictions, producing
segmentation masks that more closely follow the ground-truth boundaries.
These examples are intended to illustrate the qualitative behavior of the
method; aggregate performance is reported using the complete test sets in
the main paper.

\section{Positional-Language Analysis}
\label{app:position-analysis}

For every report, we constructed four deterministic interventions: masking the position phrase,
replacing it with a location-agnostic phrase, removing it, and masking an
equal number of non-position tokens. The image, checkpoint, preprocessing,
and prediction threshold were fixed, so the analysis measures sensitivity
to text changes rather than training variation.

\begin{figure}[t!]
\centering
\includegraphics[width=\columnwidth]
{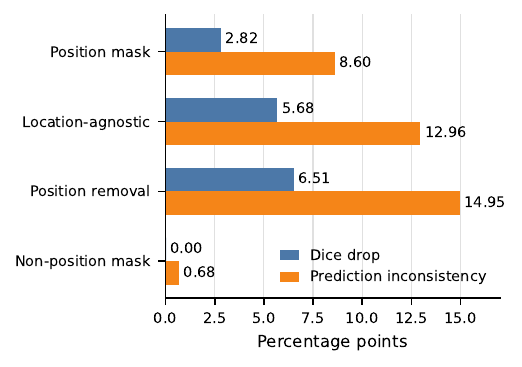}
\caption{Cohort-level effects of controlled text interventions on the
QaTa-COV19 validation set. The original-report mean per-image segmentation
Dice is 68.21\%. Dice drop is the decrease in mean per-image Dice against
the ground-truth masks. Prediction inconsistency is $100$ minus the mean
Dice overlap between predictions obtained with the original and intervened
reports. Stronger degradation of positional content produces larger changes under both measures, whereas matched non-position masking has negligible effect. All values are percentages.}
\label{fig:position-intervention-summary}
\end{figure}

Fig.~\ref{fig:position-intervention-summary} reports the corresponding
per-image results. Dice is computed against the ground-truth mask for each
image and then averaged, so these values are not directly comparable with
the dataset-level Dice in the main tables. Complete position removal causes the largest degradation, whereas masking the same number of non-position
tokens has negligible effect. Separately, Fig.~1(a) of the main paper
provides a qualitative laterality diagnostic in which swapping left/right
redirects the predicted region.

\paragraph{Attention summary.}
The token analysis uses valid, non-padding report tokens and identifies
laterality and vertical-location tokens with the same position lexicon used
by PosAug. Attention mass is normalized within each fusion stage before
aggregation. The summary in Fig.~1(b) of the main paper is
descriptive and is not used for model selection, hyperparameter tuning, or
training supervision.

\section{PosMed Cross-Domain Protocol}
\label{app:posmed}

Each original PosMed record contains a medical image, a binary target mask, a
free-form question, a natural-language answer, spatial metadata, and a split
identifier. Its original position-reasoning formulation takes the image and
question as input and jointly produces the answer and segmentation mask. We
reformulate it for MRIS as $(I,A)\rightarrow Y$: the original answer $A$ is
used as the referring expression and the associated binary mask $Y$ is the
segmentation target. Neither the question nor the dedicated position column is
provided to the model. Spatial tags needed by PosAug, T-PatchMix, and PACL are
parsed from the answer only; parsing succeeds for 38,725 of the 38,726 retained
records. The one unparsed record remains usable for segmentation but does not
contribute a position-derived target.

The original release aggregates six imaging groups: contrast-enhanced brain
MRI, breast ultrasound, lung CT, chest radiography, RGB polyp endoscopy, and
RGB skin dermoscopy. The four domains used in our cross-domain experiment come
from the Cheng Brain MRI dataset; BUSI; the combined Kvasir-SEG, ClinicDB,
CVC-300, ColonDB, and ETIS polyp collections; and ISIC, respectively. Thus, the
experiment
spans MRI, ultrasound, endoscopic RGB imaging, and dermoscopic RGB imaging;
the two lung groups are excluded because the main paper already evaluates
chest CT and radiography datasets.

The preprocessing artifact covers all six native PosMed subsets. Table~6
of the main paper reports the four non-lung lesion-centric subsets---brain
MRI, breast ultrasound, polyp endoscopy, and skin dermoscopy---to evaluate
transfer beyond
the chest-image benchmarks in the main comparison. Table~\ref{tab:posmed-split}
lists the exact records from these four reported domains. A single model is
optimized on the unified training stream with modality-balanced sampling;
validation macro Dice across domains selects the checkpoint. Images and masks
are resized to $224\times224$, answer text is truncated to 48 tokens, and the
remaining Semi-MedRef settings follow Table~\ref{tab:final-hyperparameters}.

\begin{table*}[t!]
\centering
\small
\setlength{\tabcolsep}{7pt}
\begin{tabular}{lrrrrr}
\toprule
Domain & Train & Labeled & Unlabeled & Validation & Test \\
\midrule
Brain MRI & 1,843 & 276 & 1,567 & 679 & 542 \\
Breast ultrasound & 599 & 90 & 509 & 50 & 113 \\
Polyp endoscopy & 1,305 & 196 & 1,109 & 145 & 798 \\
Skin dermoscopy & 806 & 121 & 685 & 90 & 379 \\
\bottomrule
\end{tabular}
\caption{PosMed split used for the four non-lung domains reported in the main
paper. Labeled and unlabeled counts partition the training column at the 15\%
mask-label budget.}
\label{tab:posmed-split}
\end{table*}

\begin{table}[t!]
\centering
{\footnotesize
\setlength{\tabcolsep}{1.5pt}
\begin{tabular}{@{}lccccc@{}}
\toprule
Method
& \makecell{Brain\\MRI}
& \makecell{Breast\\US}
& \makecell{Polyp\\Endoscopy}
& \makecell{Skin\\Dermoscopy} \\
\midrule
TeViD & 62.26/45.20 & 70.03/53.88 & 72.47/56.83 & 87.93/78.46 \\
TextMoE & 63.60/46.62 & 82.48/70.18 & 75.99/61.28 & 90.60/82.82 \\
Semi-MedRef & \textbf{67.00/50.37} & \textbf{84.90/73.76}
& \textbf{77.54/63.31} & \textbf{91.92/85.05} \\
\bottomrule
\end{tabular}}
\caption{Additional semi-supervised comparison on the four non-lung PosMed
domains using the same 15\% labeled split as Table~6 of the main paper. Each
entry reports Dice/mIoU (\%). TextMoE provides an additional SSL reference to
the TeViD comparison reported in the main paper.}
\label{tab:posmed-textmoe}
\end{table}

\begin{figure*}[t]
\centering
\includegraphics[width=0.60\textwidth]{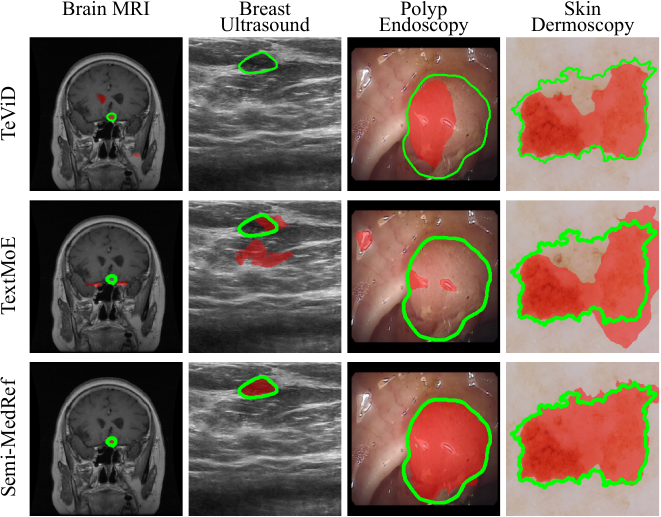}
\caption{Qualitative comparison among TeViD, TextMoE, and Semi-MedRef
under the 5\% labeled PosMed setting across brain MRI, breast ultrasound,
polyp endoscopy, and skin dermoscopy. All methods are evaluated on the
same selected samples. Red regions denote predicted masks and green
contours denote ground truth.}
\label{fig:posmed-textmoe-qualitative}
\end{figure*}

Figure~\ref{fig:posmed-textmoe-qualitative} provides an additional 5\%-label
qualitative comparison, visualizing cross-modality differences in lesion
coverage and boundary alignment; Table~\ref{tab:posmed-textmoe} remains the
aggregate comparison under the main paper's 15\%-label PosMed protocol.

The 15\% labeled subset contains the fixed 5\% subset. Sampling preserves
the highest available grouping level: patient for brain MRI, study for lung
CT, and case for the remaining subsets. Records with identical image
SHA-256 hashes are assigned to the same partition. We excluded five records
whose image content was duplicated across an official test boundary. After
this quality control, no sample identifier, image hash, or recoverable
patient/study/case group overlaps across the training, validation, and test
partitions. The released preprocessing script, fixed manifests, Brain MRI
patient/fold map, and provenance file reproduce this procedure.

\paragraph{Position representation.}
We use a dataset-specific position vocabulary. For QaTa-COV19 and
MosMedData+, the position target is the six-dimensional multi-hot vector
defined in the main paper, corresponding to the upper, middle, and lower
regions of the left and right lungs. PosMed instead uses
$\mathbf q_i^{\mathrm{PosMed}}\in\{0,1\}^{5}$, whose entries represent
top-left, top-right, bottom-left, bottom-right, and center. The vector is
deterministically extracted from the positional answer, with multiple
mentioned regions producing a multi-hot target. No $6$-to-$5$ projection
is required: PACL computes position similarity within each dataset and is
agnostic to the dimensionality of its position vocabulary. ISPG is not
enabled in the PosMed experiments.

\section{ISPG Robustness Analyses}
\label{app:ispg-analysis}

ISPG is disabled in all default comparisons and component ablations. It is
activated only for the position-degraded evaluations in Table~7 of the main
paper and in this section. ``Standard'' evaluates the default model with the
complete report. For the other routes, all explicit position phrases are
removed while the remaining report is retained: ``Zero'' supplies no
position token, ``ISPG'' supplies the image-predicted soft token, and
``Oracle'' supplies the position vector parsed from the original report.

\subsection{Training-Removal and Test-Removal Rates}

Table~\ref{tab:position-removal-matrix} separates the probability of removing
position phrases during robustness training, $\rho_{\mathrm{tr}}$, from the
test-time removal rate, $\rho_{\mathrm{te}}$. The matrix is an auxiliary 1\%
QaTa-COV19 analysis. The main ISPG configuration uses
$\rho_{\mathrm{tr}}=50\%$, selected by validation performance before the
reported test evaluation.

\begin{table}[t!]
\centering
\small
\setlength{\tabcolsep}{2.8pt}
\begin{tabular}{@{}ccccc@{}}
\toprule
\multirow{2}{*}{$\rho_{\mathrm{tr}}$} &
\multicolumn{4}{c}{$\rho_{\mathrm{te}}$} \\
\cmidrule(lr){2-5}
& 0\% & 20\% & 50\% & 100\% \\
\midrule
0\% & 84.70/73.46 & 82.18/69.75 & 77.98/63.91 & 68.97/52.63 \\
20\% & \textbf{85.01/73.92} & 84.40/73.00 & 83.68/71.94 & 82.32/69.95 \\
50\% & 84.90/73.76 & \textbf{84.81/73.62} & \textbf{84.72/73.49} & \textbf{84.54/73.22} \\
100\% & 83.38/71.49 & 83.36/71.46 & 83.36/71.46 & 83.31/71.39 \\
\bottomrule
\end{tabular}
\caption{Robustness to missing positional expressions on QaTa-COV19 under the
1\% labeled setting. Each entry reports Dice/mIoU (\%).}
\label{tab:position-removal-matrix}
\end{table}

\subsection{Position Predictor and Inference Cost}

We compared three image-only predictors using the same QaTa-COV19 training
targets and selected each checkpoint using validation mAP. The global-average-
pooling MLP provides the best accuracy--efficiency trade-off and is therefore
used by ISPG. FLOPs are reported for a single $224\times224$ input, with one MAC counted as two FLOPs. The parameter and FLOP columns in Table~\ref{tab:position-predictor} describe each complete
image-only predictor. The deployed ISPG configuration uses the selected
independent GAP--MLP predictor. Table~\ref{tab:ispg-robustness-full} additionally reports the prediction-head-only cost for reference, while the complete ISPG overhead includes its ConvNeXt-Tiny vision encoder and prediction head.

\begin{table}[t!]
\centering
\small
\setlength{\tabcolsep}{2.2pt}
\begin{tabular}{lccccc}
\toprule
Predictor & mAP & AUROC & F1 & Params & GFLOPs \\
\midrule
GAP--MLP & \textbf{90.79} & \textbf{84.73} & \textbf{83.00} & \textbf{28.42M} & \textbf{8.93} \\
Six-region pooling & 87.97 & 81.32 & 82.00 & 29.90M & 8.94 \\
Six visual queries & 89.02 & 83.02 & 82.54 & 42.31M & 9.31 \\
\bottomrule
\end{tabular}
\caption{Accuracy--efficiency comparison of image-only position predictors.
Prediction scores are percentages; AUROC and F1 are macro-averaged.}
\label{tab:position-predictor}
\end{table}

\begin{table}[t!]
\centering
\small
\setlength{\tabcolsep}{3.0pt}
\begin{tabular}{@{}lcccc@{}}
\toprule
Labels & Standard & Zero & ISPG & Oracle \\
\midrule
\multicolumn{5}{c}{\textbf{QaTa-COV19}} \\
\midrule
2\% & 87.25/77.38 & 81.82/69.23 & 86.12/75.62 & 87.01/77.01 \\
5\% & 88.31/78.98 & 75.80/61.03 & 87.55/77.85 & 88.45/79.29 \\
15\% & 89.84/81.56 & 74.86/59.82 & 88.57/79.48 & 89.75/81.40 \\
\midrule
\multicolumn{5}{c}{\textbf{MosMedData+}} \\
\midrule
2\% & 73.06/57.56 & 72.16/56.42 & 73.23/57.77 & 73.52/58.13 \\
5\% & 74.51/59.38 & 71.91/56.14 & 73.17/57.69 & 73.24/57.78 \\
15\% & 76.42/61.84 & 74.92/59.90 & 75.05/60.07 & 75.29/60.38 \\
\midrule
\multicolumn{2}{@{}l}{\textit{Prediction-head overhead}} &
\multicolumn{3}{c@{}}{$+0.30$M Params $/$ $<0.001$ GFLOPs} \\
\multicolumn{2}{@{}l}{\textit{ISPG overhead}} &
\multicolumn{3}{c@{}}{$+28.42$M Params $/$ $+8.93$ GFLOPs} \\
\bottomrule
\end{tabular}
\caption{Robustness under complete test-time removal of positional
expressions. Each performance entry reports Dice/mIoU (\%). The final two
rows report the prediction-head-only cost and the complete ISPG predictor
cost, respectively.}
\label{tab:ispg-robustness-full}
\end{table}

\section{Statistical Significance Analysis}
\label{app:statistical-significance}

\subsection{Analysis Family and Statistical Unit}

As an ancillary checkpoint-level analysis distinct from the three-seed
summary in Table~1 of the main paper,  we evaluated the final
2\%-label QaTa-COV19 comparisons with MMI-UNet and GuideDecoder. Testing Dice and mIoU produced a family of four hypotheses. Predictions were paired on the same 2,113 QaTa-COV19 test images, which
were grouped into 474 patient clusters. All resampling and randomization
operations retained complete patient clusters, preventing images from the
same patient from being treated as independent observations. Test labels
were used only for final evaluation and did not affect training or
checkpoint selection.

\subsection{Metrics and Effect Estimate}
\label{app:metric-definitions}

Final evaluation used the student checkpoint selected by validation Dice and
the fixed test manifest. Foreground probabilities were thresholded at 0.5.
Dataset-level metrics were computed by summing foreground true-positive
($TP$), false-positive ($FP$), and false-negative ($FN$) counts over the
complete test cohort:
\begin{equation}
\begin{aligned}
\operatorname{Dice} &=
\frac{2TP}{2TP+FP+FN},\\
\operatorname{mIoU} &=
\frac{TP}{TP+FP+FN}.
\end{aligned}
\end{equation}
Thus, the reported values are global dataset-level scores rather than
unweighted means of per-image metrics. Dice measures foreground overlap under
class imbalance, while mIoU provides a stricter union-normalized measure and
follows standard MRIS evaluation practice.

The paired effect estimate is
\begin{equation}
\Delta_M = M_{\text{Semi-MedRef}} - M_{\text{baseline}},
\end{equation}
where $M$ denotes Dice or mIoU. Differences are reported in percentage points
(pp); for example, $+2.62$ denotes an increase from 84.63\% to 87.25\%.

\subsection{Confidence Intervals and Hypothesis Tests}
We obtained 95\% percentile confidence intervals with 50,000 paired
patient-cluster bootstrap repetitions. In each repetition, patient clusters
were sampled with replacement, and all images belonging to each sampled
patient were retained for both models before recomputing the dataset-level
metric and paired difference.

The primary significance test was a two-sided paired patient-cluster
randomization test. Under the null hypothesis of exchangeability, the
complete prediction sets of the baseline and Semi-MedRef models were swapped
within each patient cluster. We used 200,000 Monte Carlo patient-cluster
swaps and calculated:
\begin{equation}
p =
\frac{1+\sum_{b=1}^{B}
\mathbb{I}\!\left(
\lvert\Delta_M^{(b)}\rvert
\geq
\lvert\Delta_M^{\mathrm{obs}}\rvert
\right)}
{B+1},
\qquad B=200{,}000.
\end{equation}
Holm's step-down procedure controlled the family-wise error rate over the
four backbone--metric hypotheses. Statistical
decisions below are based on the Holm-adjusted randomization $p$ values; the
bootstrap intervals are unadjusted descriptive uncertainty intervals.

\begin{table}[t!]
\centering
\footnotesize
\setlength{\tabcolsep}{3pt}
\begin{tabular}{lcc}
\toprule
Comparison & $\Delta$ (95\% CI), pp & $p_{\mathrm{Holm}}$ \\
\midrule
MMI Dice & $+2.62$ [$+2.28,+2.97$]
& $\mathbf{2.00{\times}10^{-5}}$ \\
MMI mIoU & $+4.03$ [$+3.52,+4.54$]
& $\mathbf{2.00{\times}10^{-5}}$ \\
Guide Dice & $+5.63$ [$+4.97,+6.34$]
& $\mathbf{2.00{\times}10^{-5}}$ \\
Guide mIoU & $+8.26$ [$+7.34,+9.24$]
& $\mathbf{2.00{\times}10^{-5}}$ \\
\bottomrule
\end{tabular}
\caption{Paired patient-clustered statistical analysis on QaTa-COV19
with 2\% labels. Differences are dataset-level percentage points.
CI denotes the unadjusted 95\% paired cluster-bootstrap confidence
interval, and $p_{\mathrm{Holm}}$ denotes the paired
cluster-randomization $p$ value after Holm correction over four
backbone--metric hypotheses.}
\label{tab:clustered-significance}
\end{table}

\subsection{Results and Interpretation}
As shown in Table~\ref{tab:clustered-significance}, all four effect
estimates are positive, and both Dice and mIoU improvements remain
significant for MMI-UNet and GuideDecoder after Holm correction.
These results provide patient-clustered evidence that the improvement
on QaTa-COV19 is not confined to a single segmentation backbone.

The confidence intervals quantify test-cohort sampling uncertainty for
one locked checkpoint pair per comparison, rather than optimization
variability across training seeds. The latter is reported separately
through the three-seed standard deviations in Table~1 of the main paper.

\section{Released Code, Commands, and Artifacts}
\label{app:released-code}

The anonymized release contains MMI-UNet and the additional GuideDecoder
instantiation, their final YAML recipes, the fixed low-label manifests, the
PosMed preprocessing pipeline, and the optional ISPG extension.
Table~\ref{tab:released-artifacts} maps each reproducibility item to its
released artifact. All commands are executed from the package root.

For the PosMed cross-domain experiment, the release additionally provides the
fixed nested 5\% and 15\% manifests, the public Brain MRI patient/fold mapping
with provenance, and the script that reconstructs all splits while auditing
patient/study/case groups and exact image-content reuse.

\begin{center}
\centering
\footnotesize
\setlength{\tabcolsep}{3pt}
\begin{tabular}{
  >{\raggedright\arraybackslash}p{0.29\columnwidth}
  >{\raggedright\arraybackslash}p{0.61\columnwidth}}
\toprule
Purpose & Artifact \\
\midrule
Environment & \path{environment.yml}; \path{requirements.txt} \\
Final recipes & \path{configs/} (dataset--backbone YAML files) \\
Fixed partitions & \path{data_splits/QaTa/}; \path{data_splits/MosMed/} \\
PosMed preprocessing & \path{scripts/prepare_posmed_splits.py};
  \path{data_splits/PosMed/} \\
Main training & \path{train_semi_aaai.py} \\
Main evaluation & \path{evaluate_semi.py} \\
ISPG training/test & \path{train_semi_soft_position.py};
  \path{evaluate_semi_soft_position.py} \\
\bottomrule
\end{tabular}
\captionof{table}{Principal artifacts in the anonymized code release.}
\label{tab:released-artifacts}
\end{center}

For example, the 2\% QaTa-COV19 MMI-UNet experiment is launched as
\begin{quote}
\footnotesize\ttfamily
python train\_semi\_aaai.py\\
\mbox{}\quad--config configs/qata\_mmiunet.yaml\\
\mbox{}\quad--ratio 0.02 --seed 42 --device 0
\end{quote}
The config may be replaced by any of the four dataset--backbone recipes, and
the main-table ratios are 0.02, 0.05, 0.15, and 1.0. The test command requires
an explicit checkpoint:
\begin{quote}
\footnotesize\ttfamily
python evaluate\_semi.py\\
\mbox{}\quad--config configs/qata\_mmiunet.yaml\\
\mbox{}\quad--checkpoint PATH --weights student\\
\mbox{}\quad--device 0 --output-json RESULT.json
\end{quote}
Checkpoints are written to \path{outputs/checkpoints/}, logs to
\path{outputs/logs/}, and evaluation metrics to the requested JSON file. The
YAML field \path{model_save_filename} is a short run label (\path{mmiunet} or
\path{guidedecoder}); the evaluator instead receives a checkpoint path through
its command-line argument.

\end{document}